\documentclass[runningheads]{llncs}
\usepackage{graphicx}
\usepackage{multirow,amssymb,amsmath,mathrsfs}
\DeclareMathOperator*{\argmin}{argmin}
\usepackage{algorithm}
\usepackage[noend]{algpseudocode}

\makeatletter
\def\BState{\State\hskip-\ALG@thistlm}
\makeatother

\begin{document}
\title{New Properties of the Data Distillation Method When Working With Tabular Data}
\titlerunning{Tabular Data Distillation}
%
\author{Dmitry Medvedev, Alexander D'yakonov}
%
\authorrunning{D. Medvedev and A. D'yakonov}
%
\institute{Lomonosov Moscow State University, Moscow, Russia} %
\maketitle              
\begin{abstract}
Data distillation is the problem of reducing the volume of training data while keeping only the necessary information. With this paper, we deeper explore the new data distillation algorithm, previously designed for image data. Our experiments with tabular data show that the model trained on distilled samples can outperform the model trained on the original dataset. One of the problems of the considered algorithm is that produced data has poor generalization on models with different hyperparameters. We show that using multiple architectures during distillation can help overcome this problem.

\end{abstract}
\section{Introduction}
Data distillation is an aggregation of all possible information from the original training dataset to reduce its volume. The algorithm proposed in~\cite{l1} tries to produce a small synthetic dataset, which can be used to train models reaching the same quality as with training on the original dataset. In addition, the algorithm also reduces the number of optimization steps needed for training on new data, limiting this number in the objective. 

Besides pure scientific interest, research in this new area can be very helpful in practice. For example, often the solution to one problem requires many different models to be trained on the same dataset. The creation of a new synthetic dataset that allows to simultaneously reduce training time for a large number of models with different architectures and hyperparameters would be very helpful. However, the mentioned algorithm has some drawbacks. Distilled data is poorly generalized for models not involved in the distillation process. In this paper, we examine the work of the algorithm on tabular data trying to address this problem.

The rest of the work is divided into 7 sections. In Section 2 we do a short overview of related work. A detailed description of the data distillation algorithm with its complexity analysis is located in Section 3. Section 4 consists of descriptions of architectures and the tabular dataset used in the research. In Section 5 we show results of experiments and examine the properties of synthetic tabular data. In Section 6 we examine the possibility of training models with different architectures on one synthetic dataset. Finally, we present our conclusions in Section 7.

\section{Related Work}
The basis of the data distillation algorithm is the optimization of synthetic data and learning rates with hypergradients. The application of backpropagation~\cite{l9} for optimization of hyperparameters was proposed in~\cite{l7} and~\cite{l8}. Backpropagation through L-BFGS~\cite{l11} and SGD with momentum~\cite{l12} was presented in~\cite{l10}, and more memory-efficient algorithm was proposed in~\cite{l4}. In addition,~\cite{l4} conducted experiments with data optimization.

The algorithm examined in our work was developed in~\cite{l1}, where successful distillation of the MNIST dataset~\cite{l2} was shown. Leaving only 10 examples (one for each class), and thus reducing the dataset volume by 600 times, they were able to train the LeNet model~\cite{l3} to quality close to the quality of training on the original dataset. Also, they proposed to use fixed distribution for network initialization to increase distilled data generalization, but still couldn't reach quality obtained with fixed initialization. It is important to note that authors of the considering algorithm~\cite{l1} were inspired by network distillation~\cite{l13}, that is, the transfer of knowledge from an ensemble of well-trained models into a single compact one.

The way to distill both objects and their labels was shown in~\cite{l6}. Authors showed that such distillation increases accuracy for several image classification tasks and allows distilled datasets to consist of fewer samples than the number of classes. Also, they showed the possibility to distill text data.

\section{Distillation Algorithm}
The simplest version of the algorithm developed in~\cite{l1} is the one-step version. Let $\theta_0$ be the initial model's weights vector sampled from a fixed distribution $p(\theta_0)$, $x$ be the original data, $\tilde{x}$ be the synthetic data (randomly initialized vectors), $\tilde{\eta}$ be a synthetic learning rate (positive scalar needed in the gradient descent method), and $l(x, \theta)$ be a loss function. If we do gradient descent step and get updated weights $\theta_1$ then the distillation algorithm can be written in the form of the following optimization problem:
\begin{eqnarray}
\tilde{x}^*, \tilde{\eta}^* = 
\argmin\limits_{\tilde{x}, \tilde{\eta}} \mathbb{E}_{\theta_0 \sim p(\theta_0)} l(x, \theta_1) 
= \argmin\limits_{\tilde{x}, \tilde{\eta}} \mathbb{E}_{\theta_0 \sim p(\theta_0)} l(x, \theta_0 
- \tilde{\eta} \nabla_{\theta_0} l(\tilde{x}, \theta_0)).
\end{eqnarray}
To launch the gradient descent method and find the optimum of this problem we have to calculate second-order derivatives, which is possible for the majority of loss functions and model's architectures. In general, when we want to train a model on distilled data for a few steps or even for a few epochs, the algorithm looks a bit more complicated. To describe it we introduce the concepts of external and internal steps and epochs, and the concept of internal models. So, at each internal step of each internal epoch, several internal models are trained on synthetic data. In~\cite{l1}, internal models have the same architecture and different initializations $\theta_0^{(j)} \sim p (\theta_0)$. At each external step of each external epoch, the loss function of these trained models is evaluated on the original data. After this, we can calculate the direction to make the descent step and optimize the synthetic data and learning rates. As a result, the general-case algorithm at each external step solves the following optimization problem:
\begin{eqnarray} \label{eq:2}
\begin{cases} 
\theta_0^{(j)} \sim p(\theta_0); \quad j = 1, ..., m; \\
\theta_{k+1}^{(j)} = \theta_{k}^{(j)} - \tilde{\eta}_k \nabla_{\theta} l(\tilde{x}_{i(k)}, \theta_k^{(j)}); 
\quad k = 0, ..., n-1; 
\quad i(k) = k \operatorname{mod} s;\\ 
\mathcal{L} = \frac{1}{m} \sum\limits_{j = 1}^m l(x, \theta_n^{(j)}) \rightarrow \min\limits_{\tilde{x}, \tilde{\eta}}.
\end{cases}
\end{eqnarray}

\noindent In~(\ref{eq:2}) $s$ is the number of internal steps of one internal epoch; $n$ is the total number of steps of the internal loop and $m$ is the number of internal models. Optimization requires an estimation of gradients $\nabla_{\tilde{x}} \mathcal{L}$ and $\nabla_{\tilde{\eta}} \mathcal{L}$:

\begin{gather*}
d \mathcal{L} = \sum\limits_{j = 1}^m \frac{\partial\mathcal{L}}{\partial\theta^{(j)}_n} d \theta^{(j)}_n
= \sum\limits_{j = 1}^m \frac{\partial\mathcal{L}}{\partial\theta^{(j)}_n} d\big( \theta^{(j)}_{n-1} - \tilde{\eta}_{n-1} \nabla_{\theta} l(\tilde{x}_{i(n-1)}, \theta^{(j)}_{n-1})\big) =\\
= \{g^{(j)}_{n-1} := \tilde{\eta}_{n-1} \nabla_{\theta} l(\tilde{x}_{i(n-1)}, \theta^{(j)}_{n-1})\}
= \sum\limits_{j = 1}^m
\Bigg[
\left( 
	\frac{\partial\mathcal{L}}{\partial\theta^{(j)}_n}
	-\frac{\partial\mathcal{L}}{\partial\theta^{(j)}_n}
	\frac{\partial g^{(j)}_{n-1}}{\partial\theta^{(j)}_{n-1}}
\right) d\theta^{(j)}_{n-1}  - \\
- \left( \frac{\partial\mathcal{L}}{\partial\theta^{(j)}_n} 
\frac{\partial g^{(j)}_{n-1}}{\partial\tilde{\eta}_{n-1}}\right) d\tilde{\eta}_{n-1}
- \left(\frac{\partial\mathcal{L}}{\partial\theta^{(j)}_n}
\frac{\partial g^{(j)}_{n-1}}{\partial\tilde{x}_{i(n-1)}}\right) d\tilde{x}_{i(n-1)}
\Bigg].
\end{gather*}
It is clear that if we continue to express $ \theta_{n-1}^{(j)} $ through $ \theta_{n-2}^{(j)} $ and so on we will get an expression with $ \theta_{0}^{(j)}$. After summing up all the necessary terms, we obtain the following formulas for gradients:
\begin{gather}
\nabla_{\tilde{\eta}_k} \mathcal{L} =
	\frac{\partial\mathcal{L}}{\partial\theta_{k+1}}
	\cdot
	\frac{\partial g_{k}}{\partial\tilde{\eta}_{k}};
	\quad\quad\quad
	\nabla_{\tilde{x}_j} \mathcal{L} =
	\sum\limits_{k = 0}^{n-1} I[j = i(k)]\cdot
	\frac{\partial\mathcal{L}}{\partial\theta_{k+1}}
	\cdot
	\frac{\partial g_{k}}{\partial\tilde{x}_i(k)}.
\end{gather}

\noindent Algorithms 1, 2 and 3 show the implementation of the method in pseudo-code. It's clear from the algorithm description that memory and time complexity is high, since we need to store $n$ copies of the internal model and to perform backward and forward passes through all these $n$ copies. This limitation negatively affects the performance, since the increment of $n$ significantly increases the quality of the model trained using the distilled dataset.

{\centering
\begin{minipage}{1.\linewidth}
\begin{algorithm}[H]
\caption{Main Cycle}
\begin{algorithmic}[1]
\State \textbf{Input:} $p(\theta_0)$;  $m$; $n$; $s$; $T$ --- number of external steps; batch sizes.
\State \textbf{Initialization} $\tilde{x}, \tilde{\eta}$
\Loop{ for each $t = 1,...,T$}:
\State $\nabla_{\tilde{x}} \mathcal{L} =  \mathbf{0}, \nabla_{\tilde{\eta}} \mathcal{L} = \mathbf{0}$ // accumulated values of opt. directions.
\State Get a minibatch of real training data $x$
\Loop{ for each model $j = 1, ..., m$}:
\State Model initialization $\theta^{(j)}_0 \sim p(\theta_0)$
\State Res $\leftarrow$ \textbf{Forward} // see Algorithm 2
\State $\nabla_{\tilde{x}} \mathcal{L}, \nabla_{\tilde{\eta}} \mathcal{L}$  $\leftarrow$ \textbf{Backward} // see Algorithm 3
\EndLoop
\textbf{Update} $\tilde{x}, \tilde{\eta}$
\EndLoop
\end{algorithmic}
\end{algorithm}
\end{minipage}
}

{\centering
\begin{minipage}{1.\linewidth}
\begin{algorithm}[H]
\caption{Forward Pass}
\begin{algorithmic}[1]
\State \textbf{Input:} $\theta^{(j)}_0, \tilde{x}, \tilde{\eta}, x$.
\State Res $\leftarrow \theta^{(j)}_0$
\Loop{ for each internal step $k = 0,...n-1$}:
\State $g_k = \tilde{\eta}_{k} \nabla_{\theta} l(\tilde{x}_{i(k)}, \theta^{(j)}_k)$
\State $\theta^{(j)}_{k+1} = \theta^{(j)}_k -  g_k$
\State Res $\leftarrow$ $g_k, \theta^{(j)}_{k+1}$ // remember comp. graph and model
\EndLoop
\State $\frac{\partial\mathcal{L}}{\partial\theta^{(j)}_n} = \frac{1}{m} \frac{\partial l(x, \theta^{(j)}_n)}{\partial\theta^{(j)}_n}$
\State Res $\leftarrow$ $\frac{\partial\mathcal{L}}{\partial\theta^{(j)}_n}$
\State \textbf{Output:} Res 
\end{algorithmic}
\end{algorithm}
\end{minipage}
}

{\centering
\begin{minipage}{1.\linewidth}
\begin{algorithm}[H]
\caption{Backward Pass}
\begin{algorithmic}[1]
\State \textbf{Input:} $\nabla_{\tilde{x}} \mathcal{L}, \nabla_{\tilde{\eta}} \mathcal{L}$, Res // Res --- comp. graphs and weights of models.  
\State $\frac{\partial\mathcal{L}}{\partial\theta^{(j)}_n}$ $\leftarrow$ Res
\Loop{ for each internal step $k = n-1,...0$}:
\State $g_k, \theta^{(j)}_k \leftarrow Res$
\State $\nabla_{\tilde{x}_{i(k)}}  \mathcal{L} \leftarrow \nabla_{\tilde{x}_{i(k)}}  \mathcal{L} + \frac{\partial\mathcal{L}}{\partial\theta^{(j)}_{k+1}} \frac{\partial g_{k}}{\partial\tilde{x}_{i(k)}}$
\State $\nabla_{\tilde{\eta}_{k}}  \mathcal{L} \leftarrow \nabla_{\tilde{\eta}_{k}}  \mathcal{L} + \frac{\partial\mathcal{L}}{\partial\theta^{(j)}_{k+1}} \frac{\partial g_{k}}{\partial\tilde{\eta}_{k}}$
\State  $\frac{\partial\mathcal{L}}{\partial\theta^{(j)}_{k}} = \frac{\partial\mathcal{L}}{\partial\theta^{(j)}_{k+1}} \left(1 - \frac{\partial g_{k}}{\partial\theta^{(j)}_{k}} \right)$
\EndLoop
\State \textbf{Output: $\nabla_{\tilde{x}}  \mathcal{L}, \nabla_{\tilde{\eta}}  \mathcal{L}$ } 
\end{algorithmic}
\end{algorithm}
\end{minipage}
}

\section{Data and Models}
\begin{figure}[h!]
\begin{minipage}[h]{0.5\linewidth}
\center{\includegraphics[width=.8\linewidth]{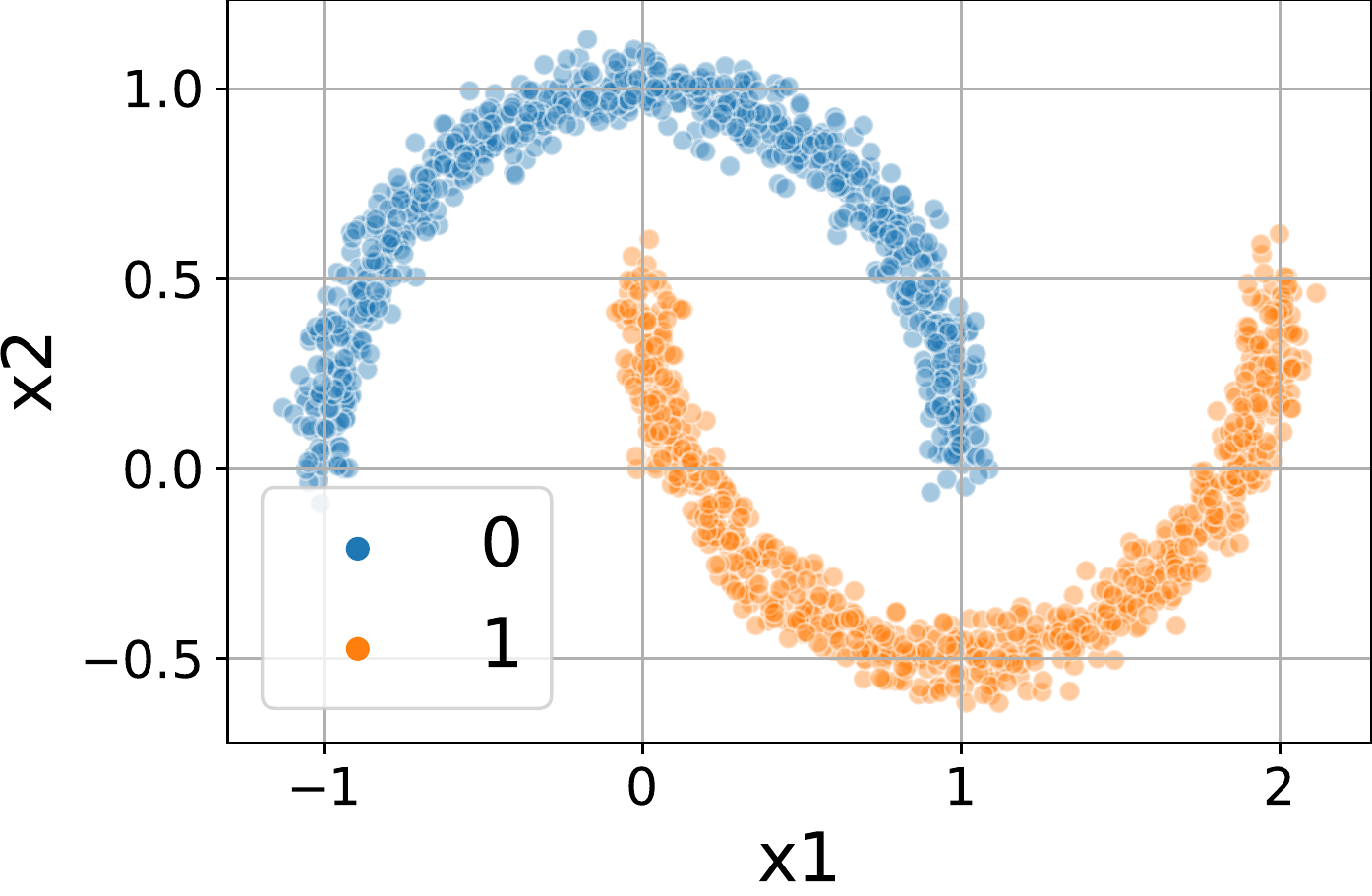}}
\end{minipage}
\hfill
\begin{minipage}[h]{0.24\linewidth}
\center{\includegraphics[width=.5\linewidth]{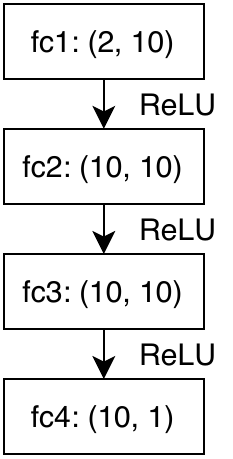}}
\end{minipage}
\hfill
\begin{minipage}[h]{0.24\linewidth}
\center{\includegraphics[width=.7\linewidth]{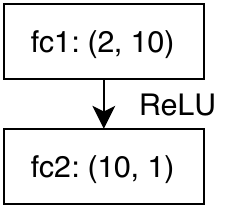}}
\end{minipage}
\hangindent=3.3cm a) \hspace{3.7cm} b) \hspace{2.45cm} c)
\caption{Used data and non-linear architectures: a) whole original dataset, b) 2-layer model, c) 4-layer model.}\label{ris1}
\end{figure}

We consider a simple two-dimensional binary classification problem, which dataset volume is 1,500 objects (see Fig.~\ref{ris1} a). The dataset is divided into training and test parts in a 2:1 ratio. We distill the training part and use the test part for quality estimation. We suppose that using such a small amount of data and producing less extreme distillation can help us to explore new properties of the algorithm. Note that it is not always possible with big visual datasets and large models due to the complexity of the distillation. For experiments we use three fully connected architectures: 1-layer, 2-layer and 4-layer. Figure~\ref{ris1} (b, c) schematically shows non-linear architectures.

\begin{figure}[h!]
\begin{minipage}[h]{0.24\linewidth}
\center{\includegraphics[width=\linewidth]{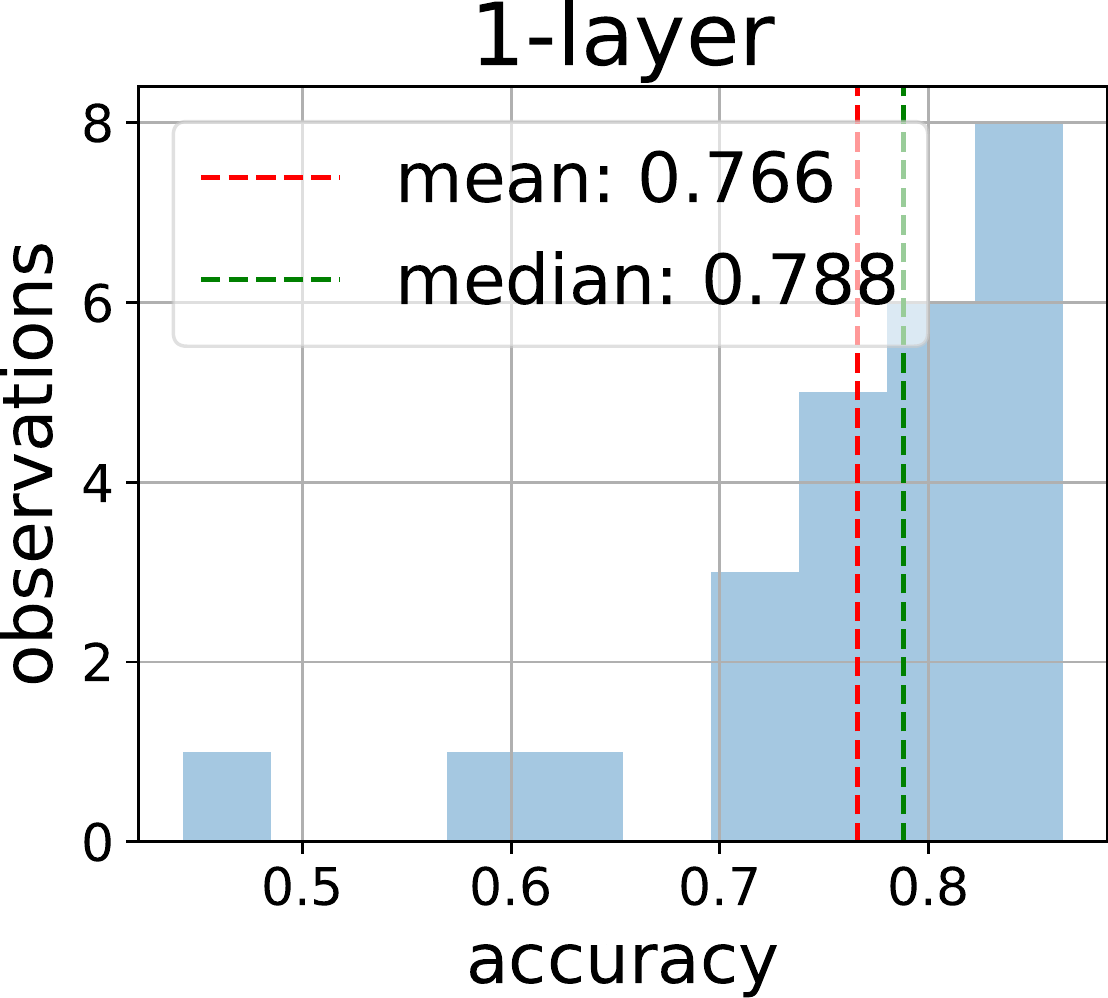}} \\a)
\end{minipage}
\hfill
\begin{minipage}[h]{0.24\linewidth}
\center{\includegraphics[width=\linewidth]{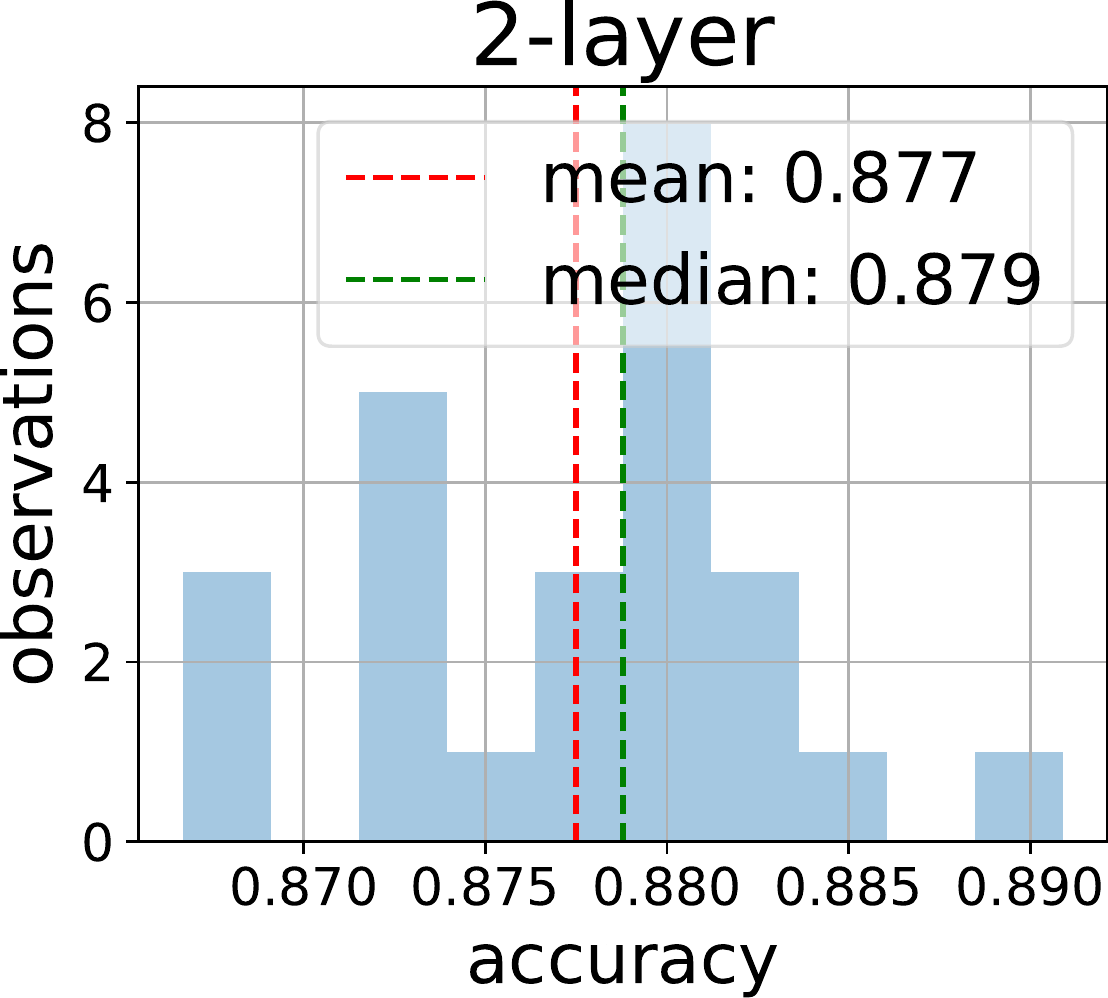}} \\b)
\end{minipage}
\hfill
\begin{minipage}[h]{0.24\linewidth}
\center{\includegraphics[width=\linewidth]{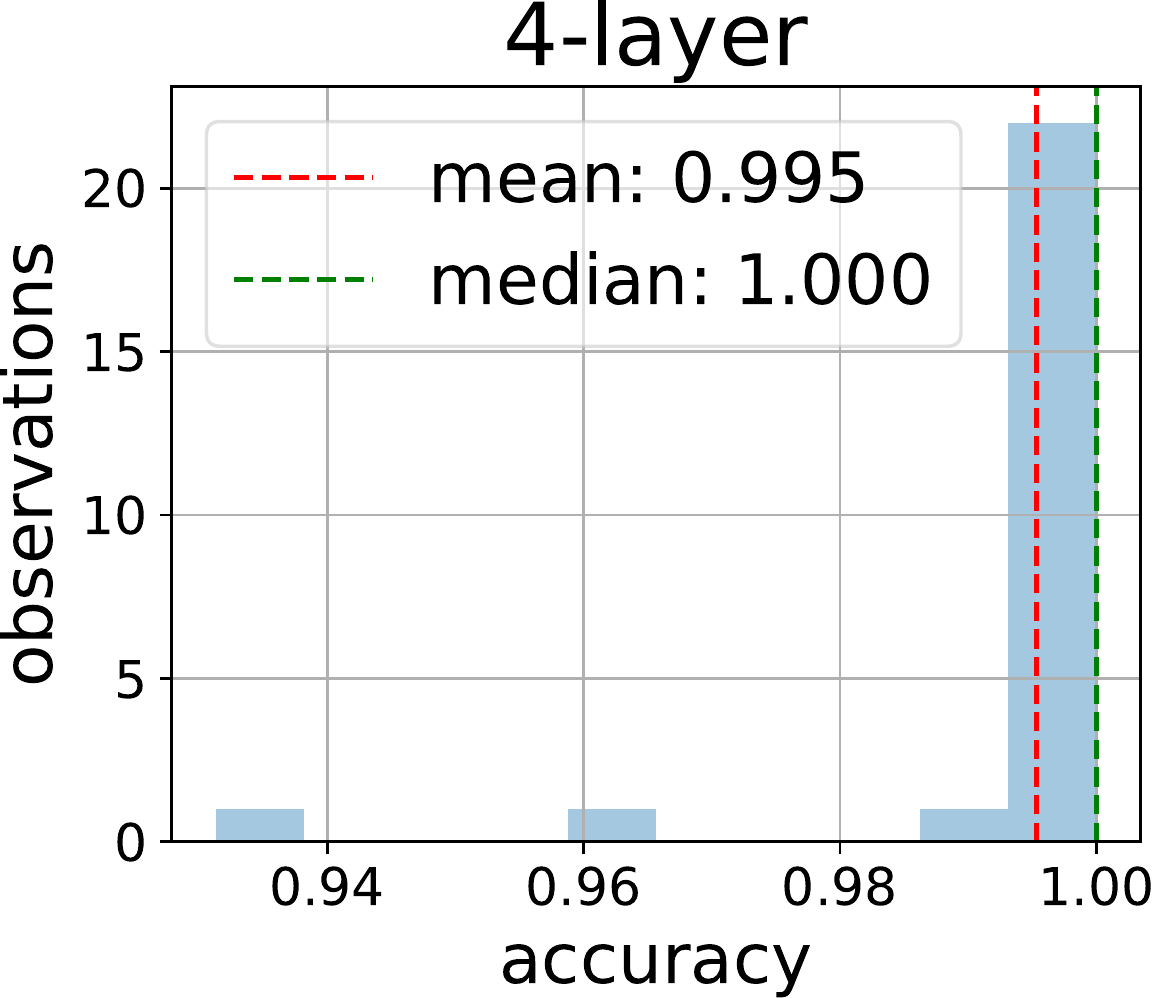}} \\c)
\end{minipage}
\begin{minipage}[h]{0.24\linewidth}
\center{\includegraphics[width=\linewidth]{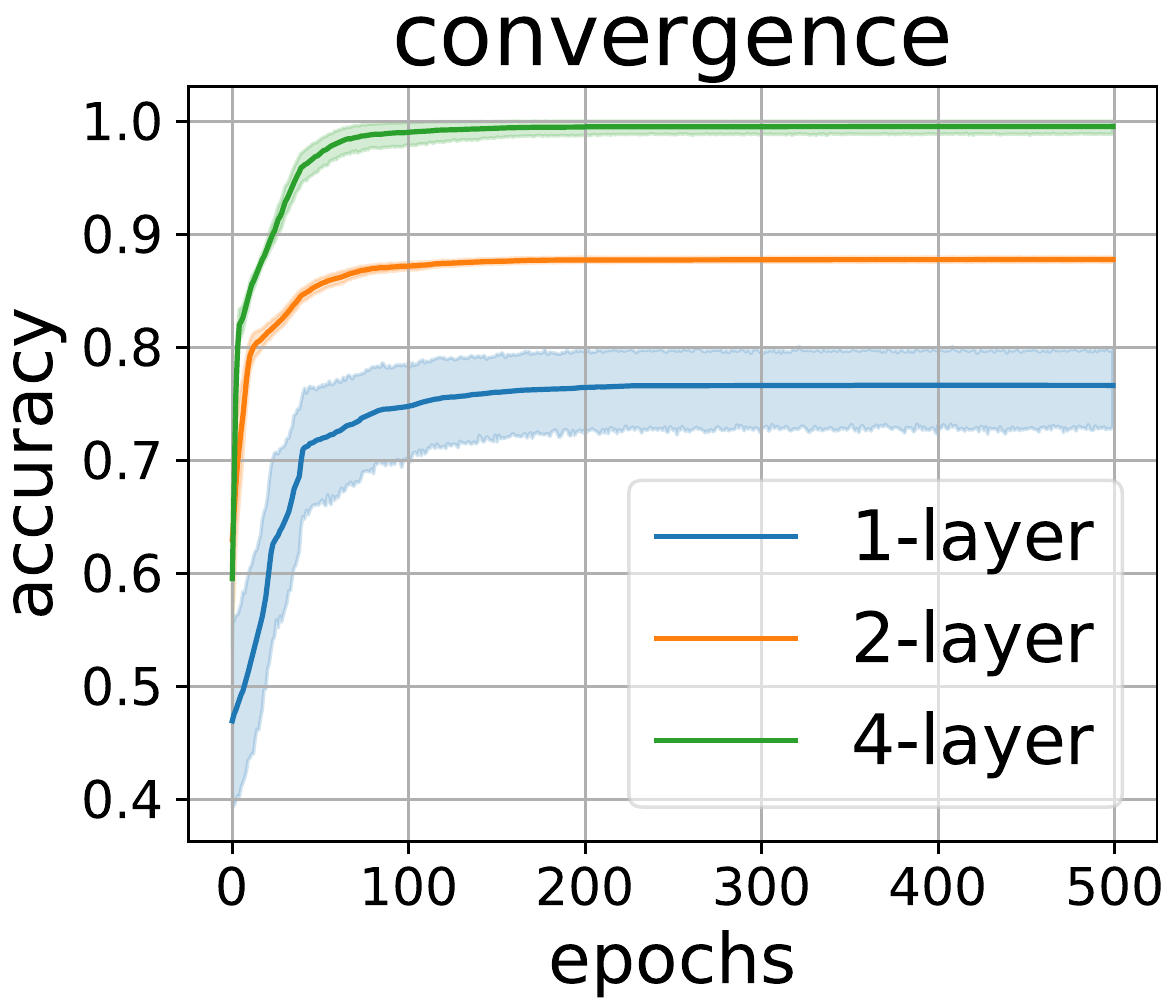}} \\d)
\end{minipage}
\caption{The quality of training on the whole dataset: a--c) distribution of accuracy for different initializations; d) the convergence of accuracy.}\label{ris2}
\end{figure}

\noindent First, we estimate the quality of training on the whole dataset. We train models 25 times with random initialization from Xavier distribution. Each training takes 500 epochs. Note that 500 epochs are more than enough for convergence for all three architectures (see Fig.~\ref{ris2} d), and it seems that the training procedure could be stopped after 200 epochs. Hereinafter, such figures show the boundaries of the 95\% confidence interval estimated with the bootstrap procedure. In addition Figure~\ref{ris2} (a-c) shows the distribution of achieved accuracy; the largest model reached the highest quality.

\begin{figure}[h!]
\begin{minipage}[h]{.24\linewidth}
\center{\includegraphics[width=1.\linewidth]{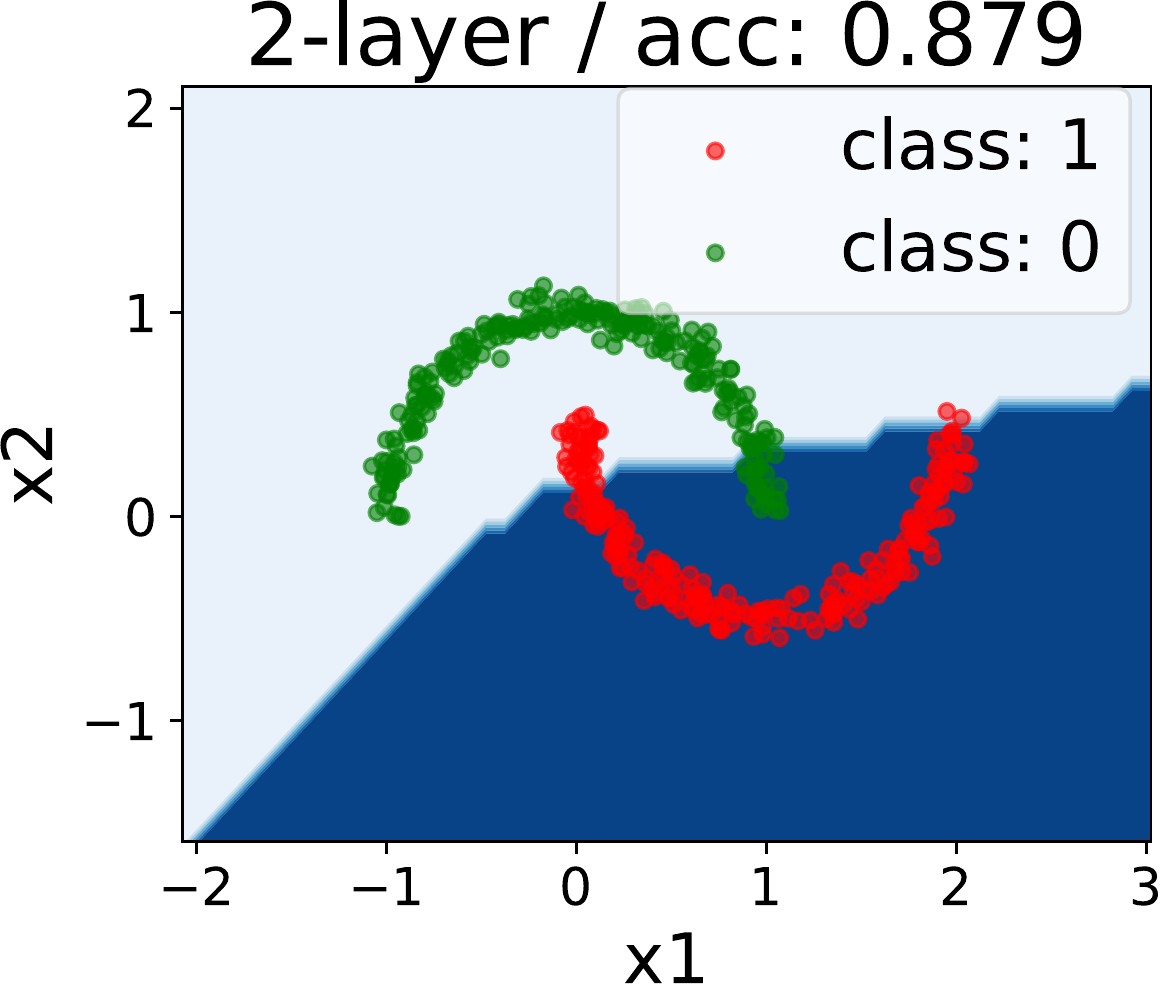}} \\a)
\end{minipage}
\hfill
\begin{minipage}[h]{.24\linewidth}
\center{\includegraphics[width=1.\linewidth]{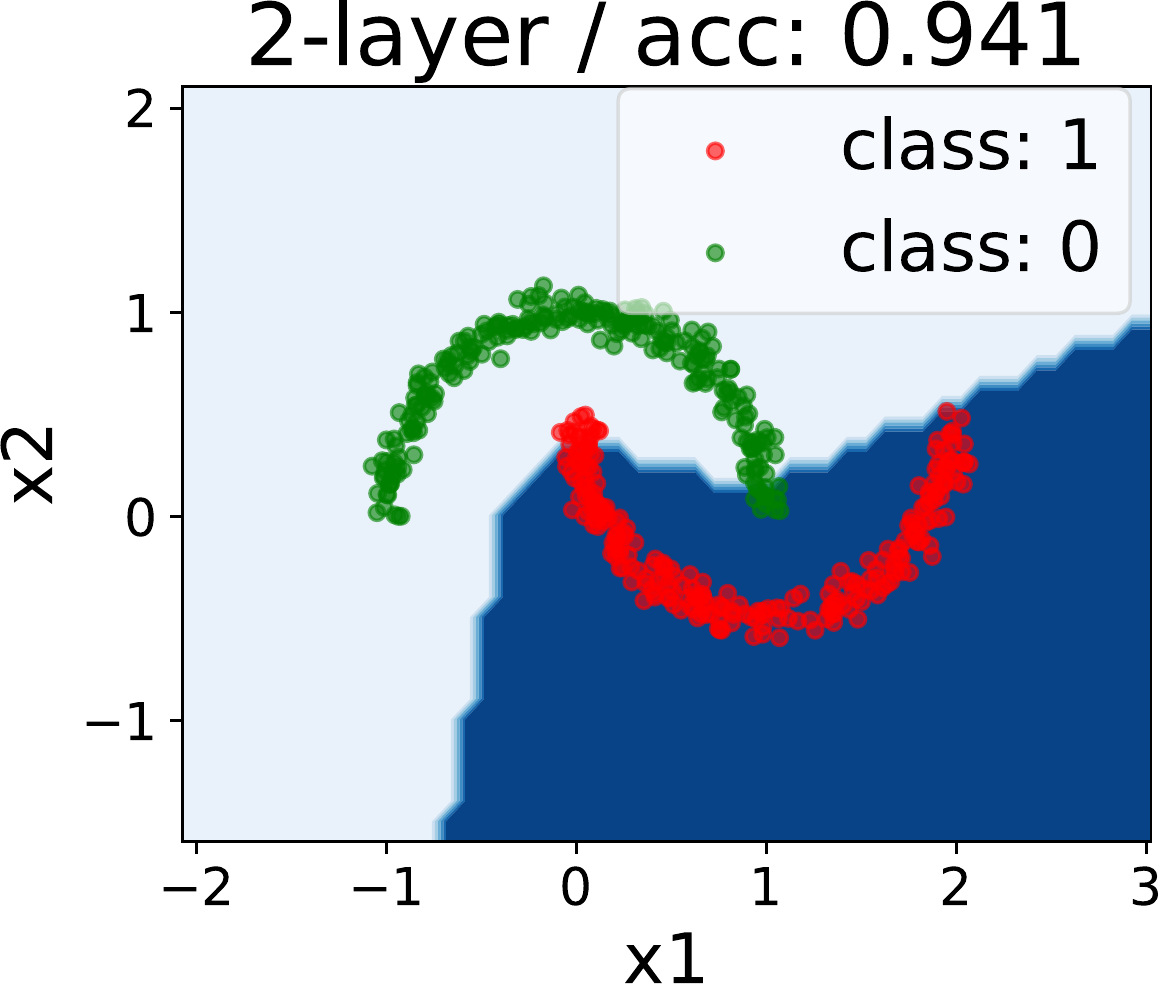}}\\b)
\end{minipage}
\hfill
\begin{minipage}[h]{.24\linewidth}
\center{\includegraphics[width=1.\linewidth]{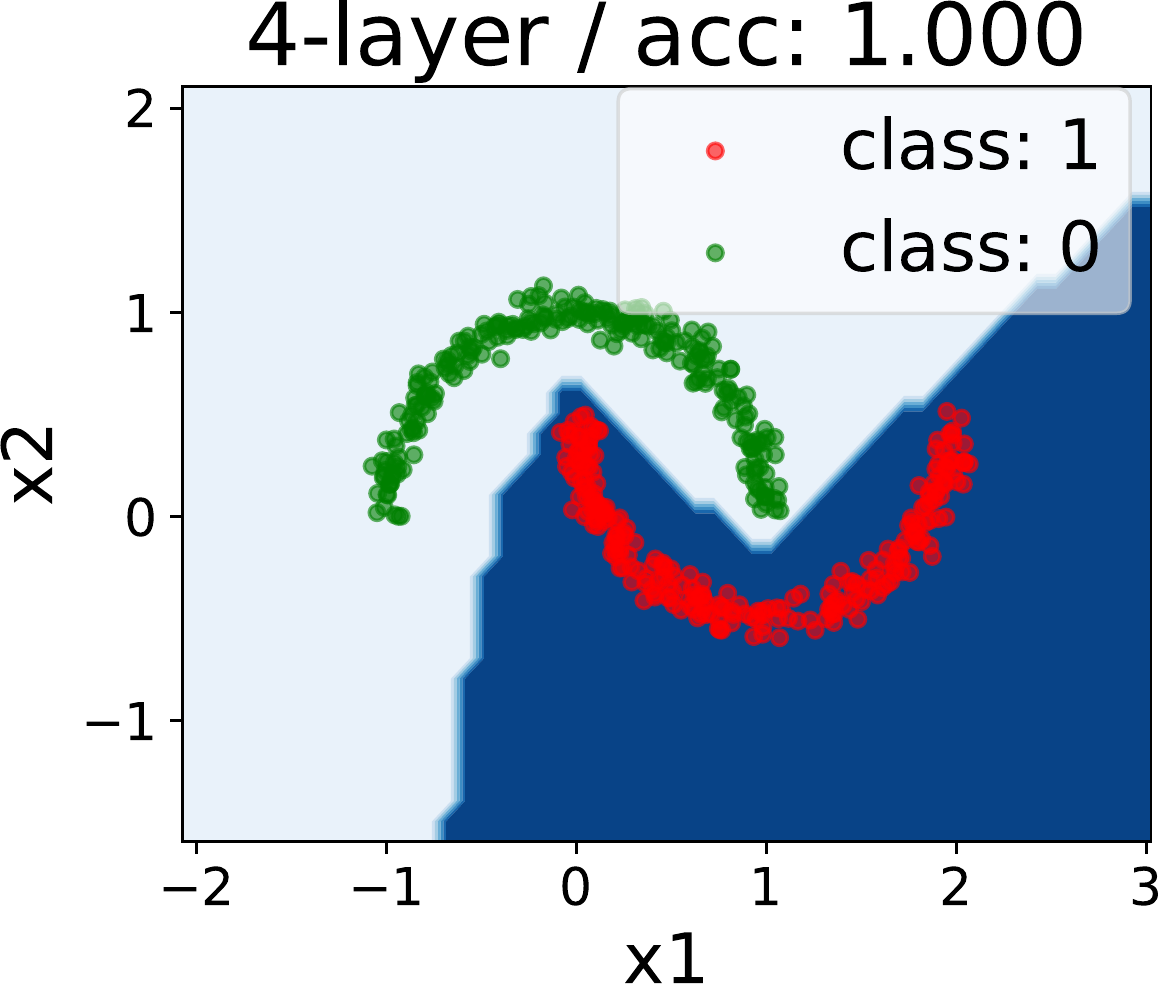}}\\c)
\end{minipage}
\hfill
\begin{minipage}[h]{.24\linewidth}
\center{\includegraphics[width=1.\linewidth]{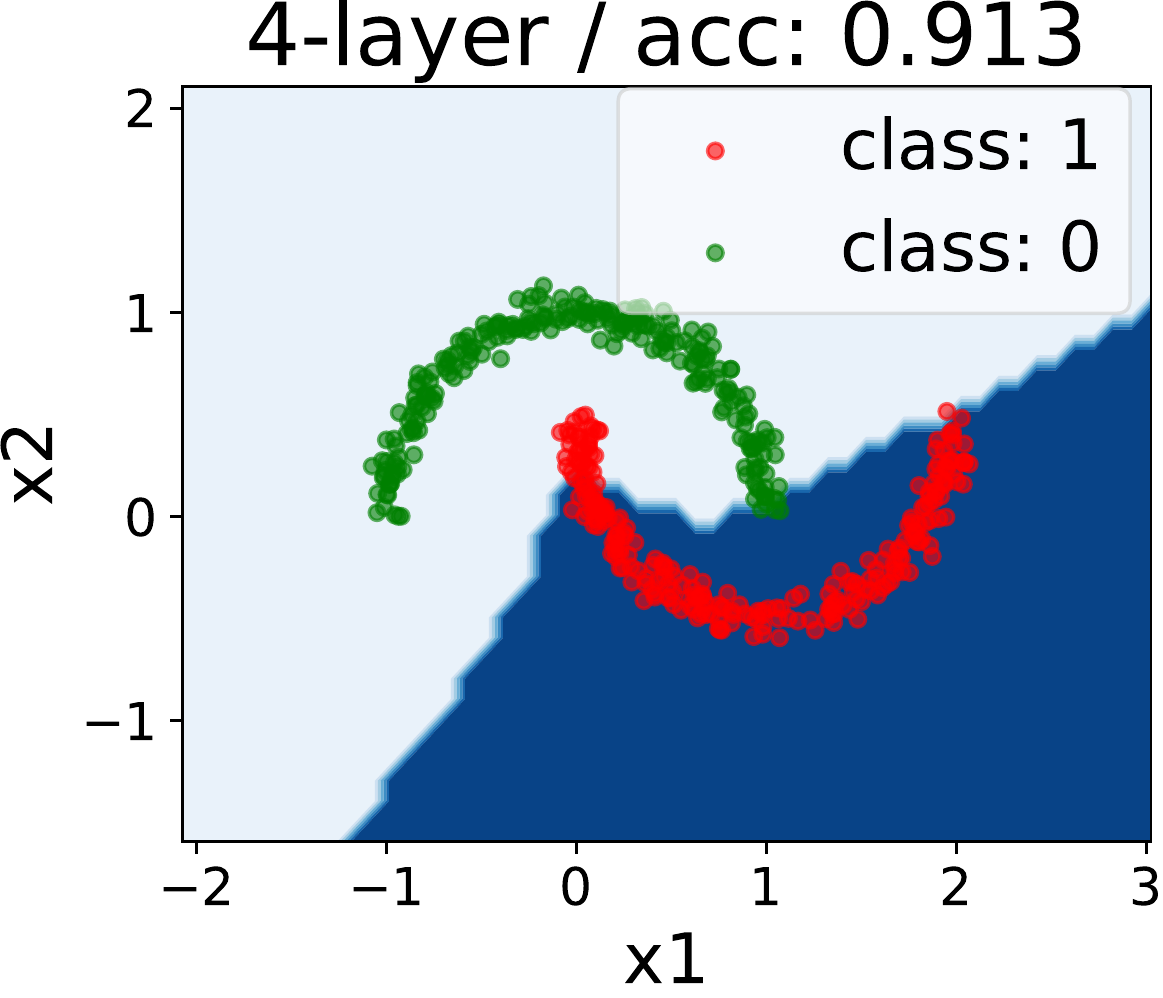}}\\d)
\end{minipage}
\caption{The decision rule boundary for models with median quality: a),c) trained on the whole dataset; b, d) trained on distilled data. The 2-layer model builds a more complex decision rule using the distilled data.}\label{ris3}
\end{figure}

\noindent Figure~\ref{ris3} (a, c) shows the boundaries of the decision rules. For each architecture, it shows the median-quality model. Note that the 2-layer model was not able to significantly outperform the linear solution, despite the significant number of made gradient descent steps and use of the whole training dataset.

\section{Tabular Data Distillation}
\subsection{Examining Hyperparameters}
The distillation algorithm has several hyperparameters: the number of internal epochs, steps and models, as they significantly affect complexity, it is important to choose the most appropriate ones. The number of internal models affects the total time of distillation. The number of steps and epochs influences $n$ and thus affects the total time of distillation and the size of needed memory. Note that the number of steps needed to train any model on distilled data is fixed. It means that there is a risk that some models may not converge in the preselected number of steps. Another parameter is the number of synthetic objects taking part in each inner step. We choose this parameter to be 4 since it seems to be enough to describe the border of two classes. 

\begin{figure}[h!]
\begin{minipage}[h]{0.32\linewidth}
\center{\includegraphics[width=1.\linewidth]{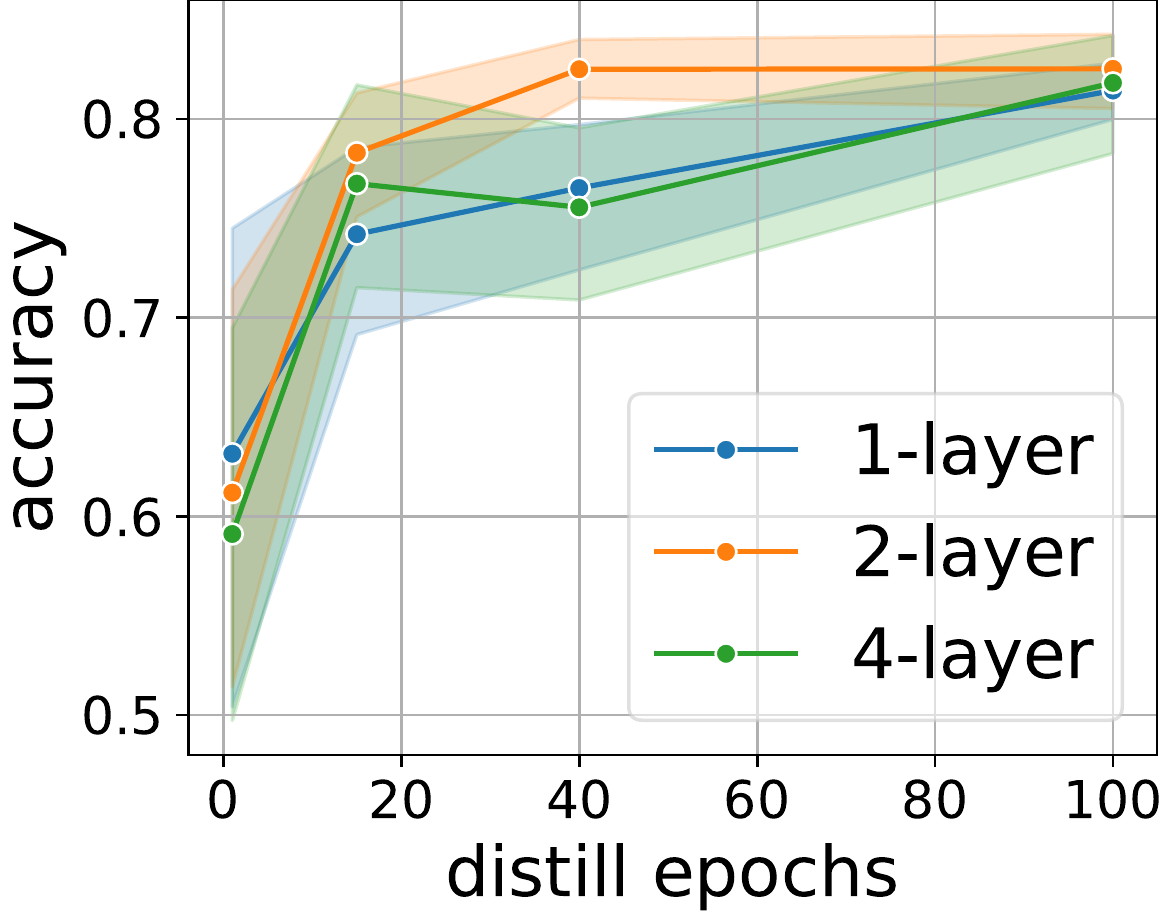}} \\a)
\end{minipage}
\hfill
\begin{minipage}[h]{0.32\linewidth}
\center{\includegraphics[width=1.\linewidth]{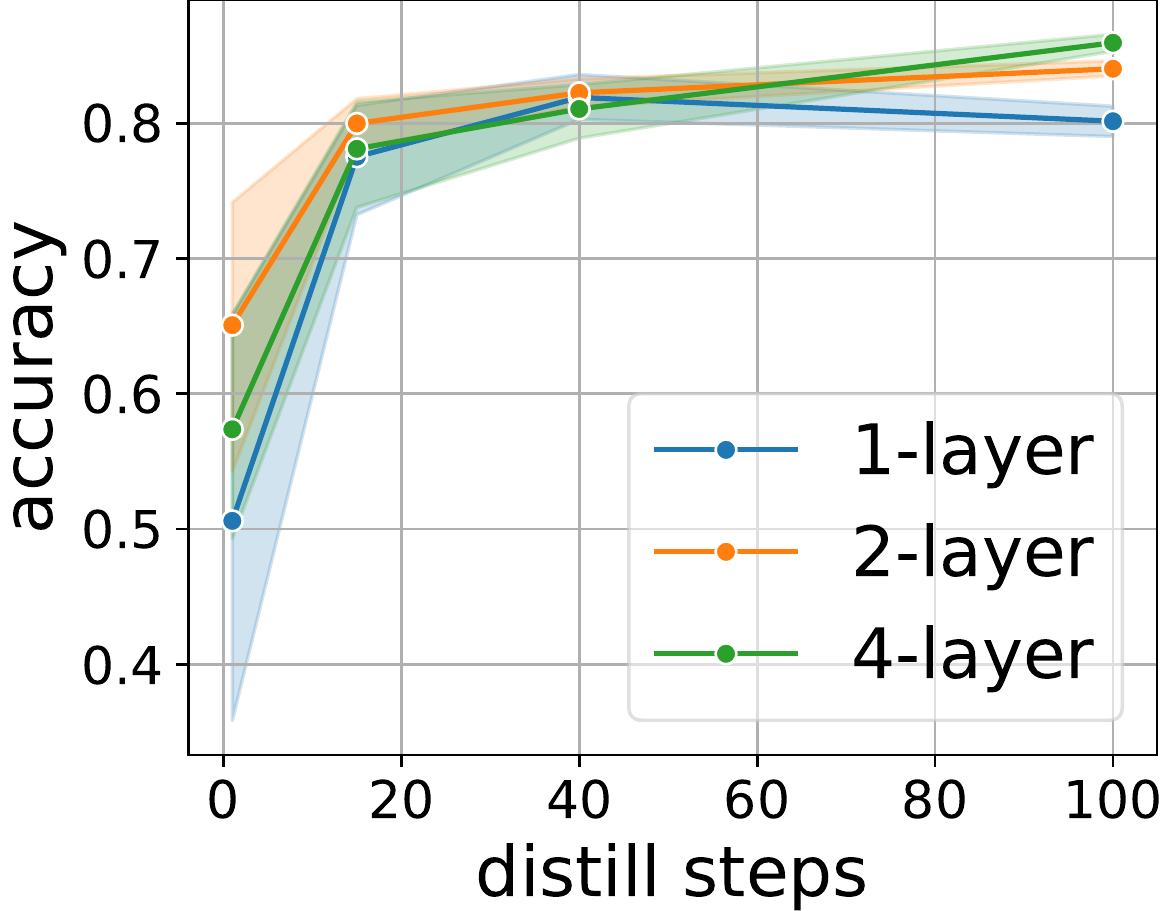}} \\b)
\end{minipage}
\hfill
\begin{minipage}[h]{0.32\linewidth}
\center{\includegraphics[width=1.\linewidth]{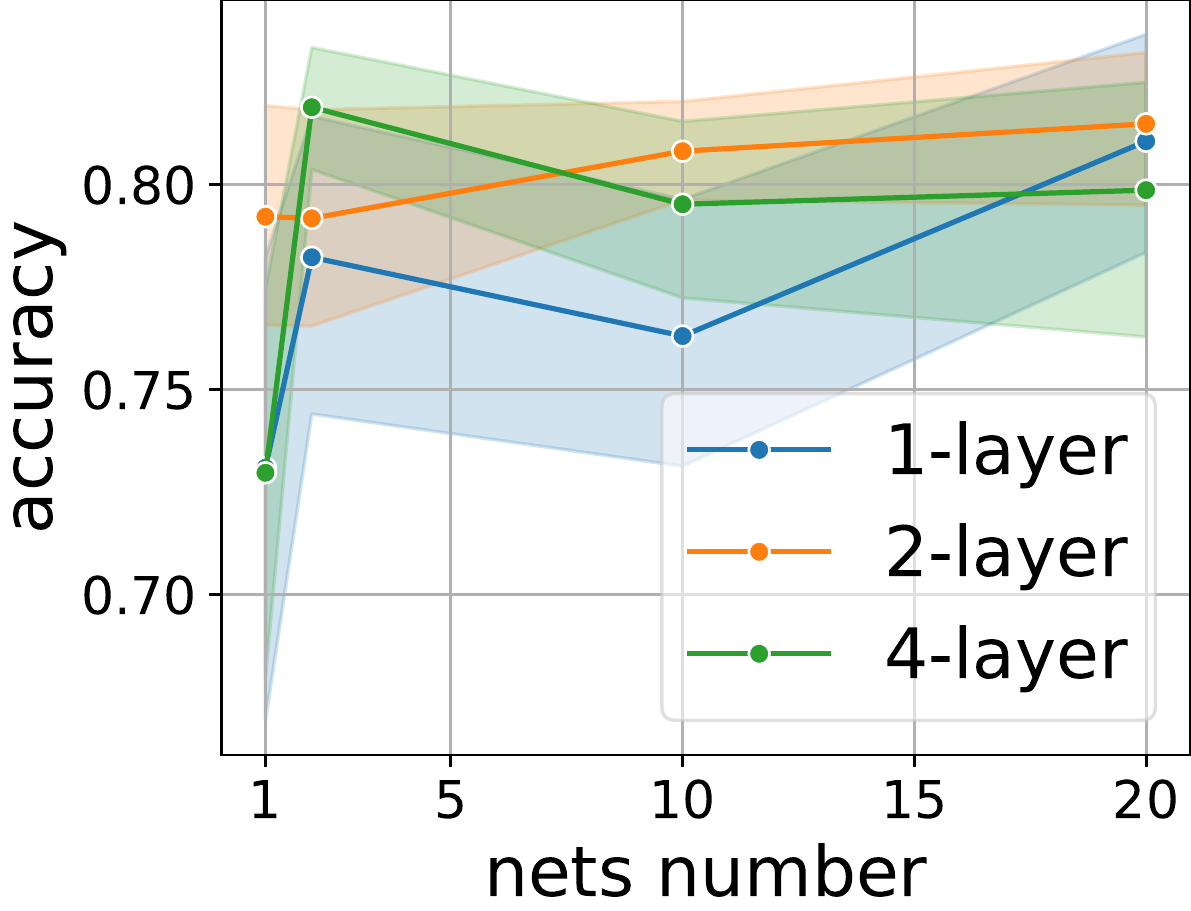}} \\c)
\end{minipage}
\caption{The dependence of the test quality on: a) the number of internal epochs (1 internal model and 1 internal step); b) the number of internal steps (1 internal model and 1 internal epoch); c) the number of internal models (10 internal steps and 1 internal epoch).}
\label{ris4}
\end{figure}

To select hyperparameters we performed several experiments similar to ones in~\cite{l1} (see Fig.~\ref{ris4}). Note that increase of both the number of epochs and steps improves the final accuracy of the algorithms, while a significant increase in the number of models does not give a noticable change.

\subsection{The Distillation Algorithm Performance}
To check the distillation algorithm performance on tabular data, we launch distillation 10 times for each architecture with different initializations of $\theta_0 $, $\tilde{x}$ and $\tilde{\eta}$). Each launch takes 50 outer epochs which is equal to 800 outer iterations since the batch size of 64. For this experiment, we choose the number of internal steps to be 40, the number of internal epochs to be 5 and the number of internal models to be 3. This set of hyperparameters leads to the total number of synthetic objects equals 320, which is more than 3 times less than the original data volume. Note that during training there are $3 \times 800 \times 5 \times 40$ forward passes through the model, and for each gradient descent step we need to store $5 \times 40$ model's copies. As a result, total distillation time for all three architectures and 10 restarts reaches 4 hours, while the usual training procedure of models on the whole original data lasts only a few minutes. To compare the standard training procedure described in Section 4 with training on distilled data we present similar plots.
\begin{figure}[h!]
\begin{minipage}[h]{0.32\linewidth}
\center{\includegraphics[width=1.\linewidth]{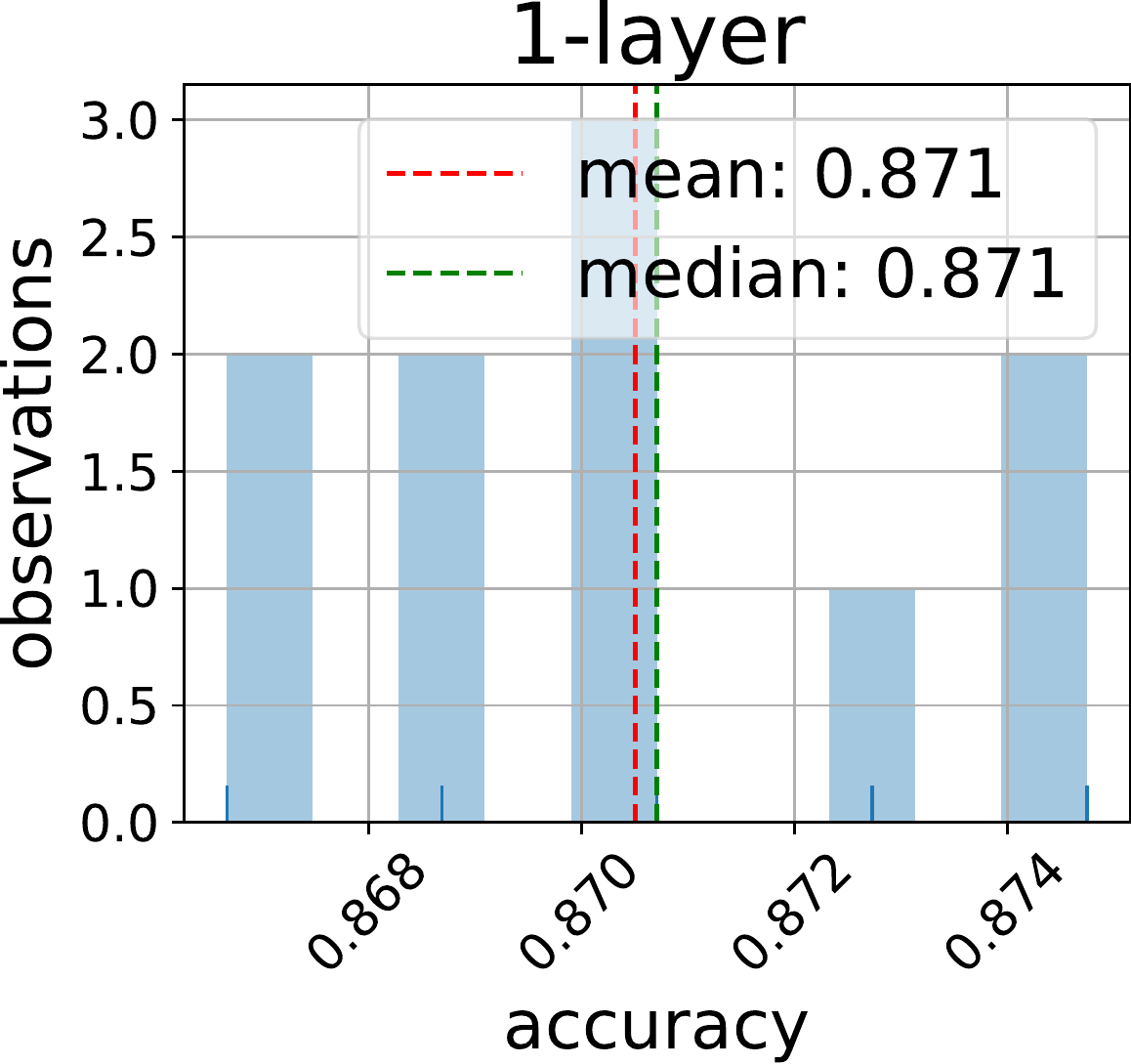}} 
\end{minipage}
\hfill
\begin{minipage}[h]{0.32\linewidth}
\center{\includegraphics[width=1.\linewidth]{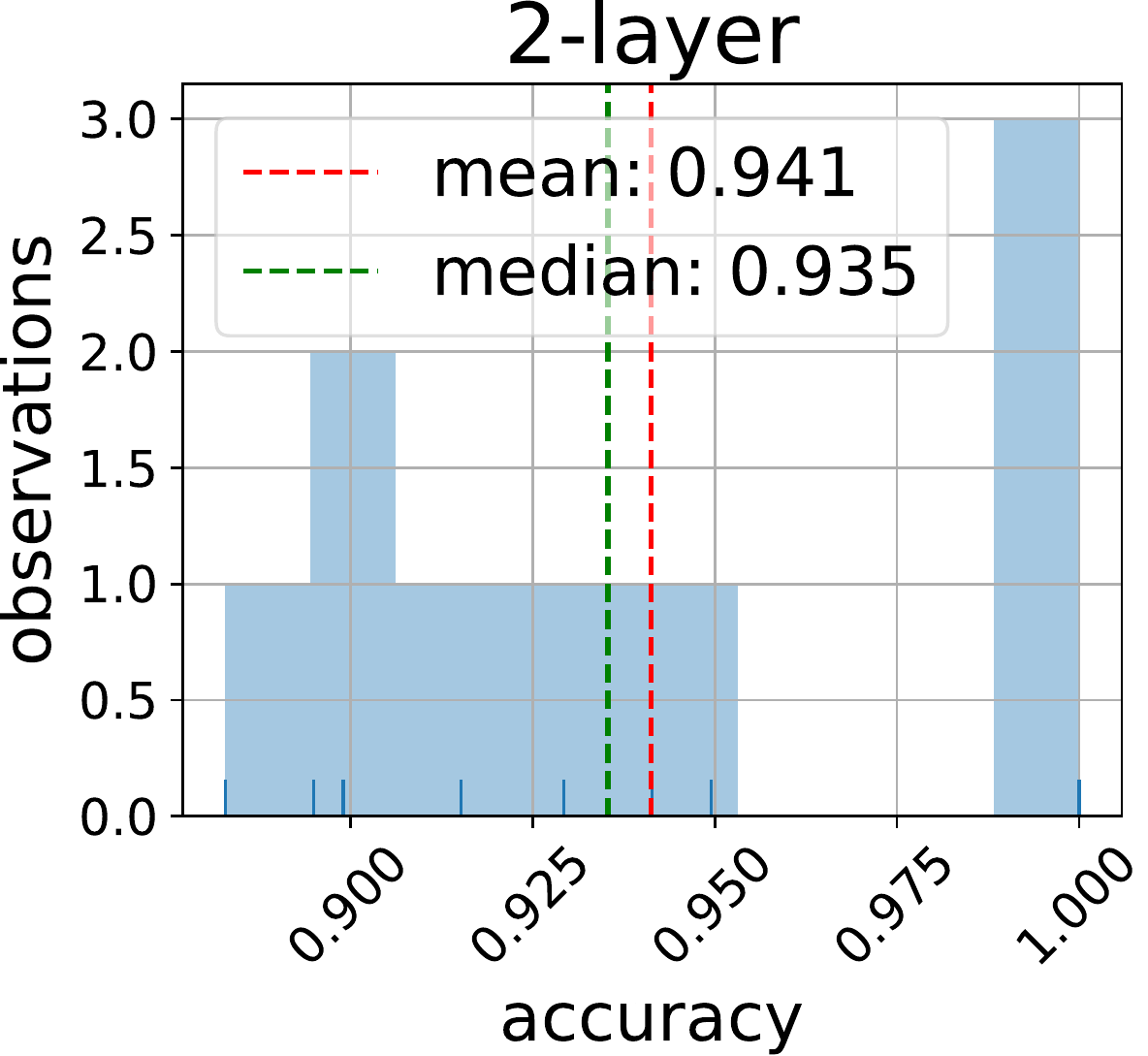}} 
\end{minipage}
\hfill
\begin{minipage}[h]{0.32\linewidth}
\center{\includegraphics[width=1.\linewidth]{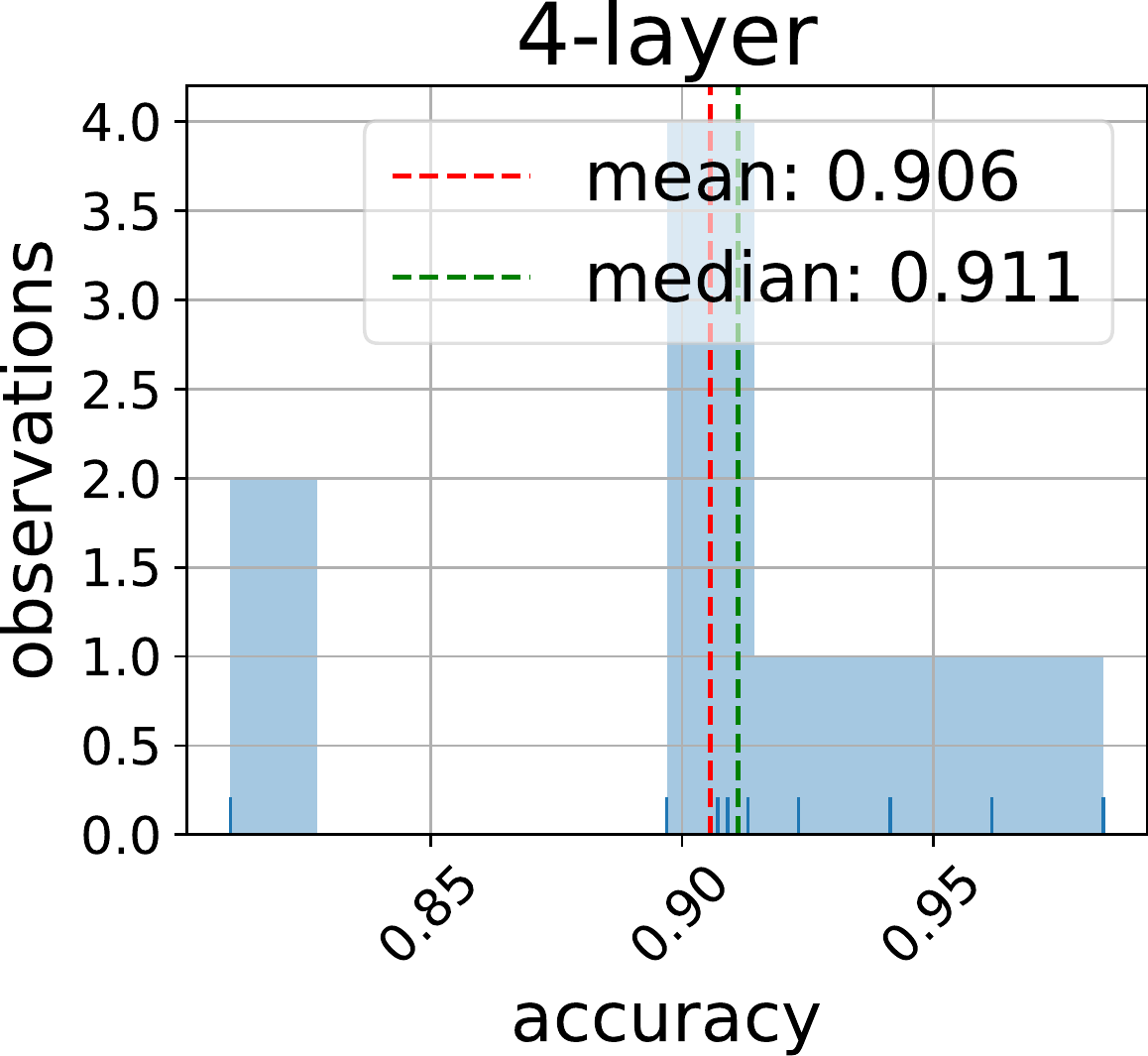}} 
\end{minipage}
\caption{Distribution of accuracy for different initializations.}
\label{ris5}
\end{figure}

\noindent Figures~\ref{ris2} (a, b, c) and~\ref{ris5} show the distribution of accuracy, and Table~\ref{table1} shows mean and standard deviation. Comparing them we can see that the quality of 1-layer and 2-layer architecture has grown.  We suggest that the distilled data has become additional parameters and allowed the network to better solve the problem. Note that at the same time, the quality of 4-layer models has decreased a bit.

\begin{figure}[h!]
\begin{minipage}[h]{.24\linewidth}
\center{\includegraphics[width=.94\linewidth]{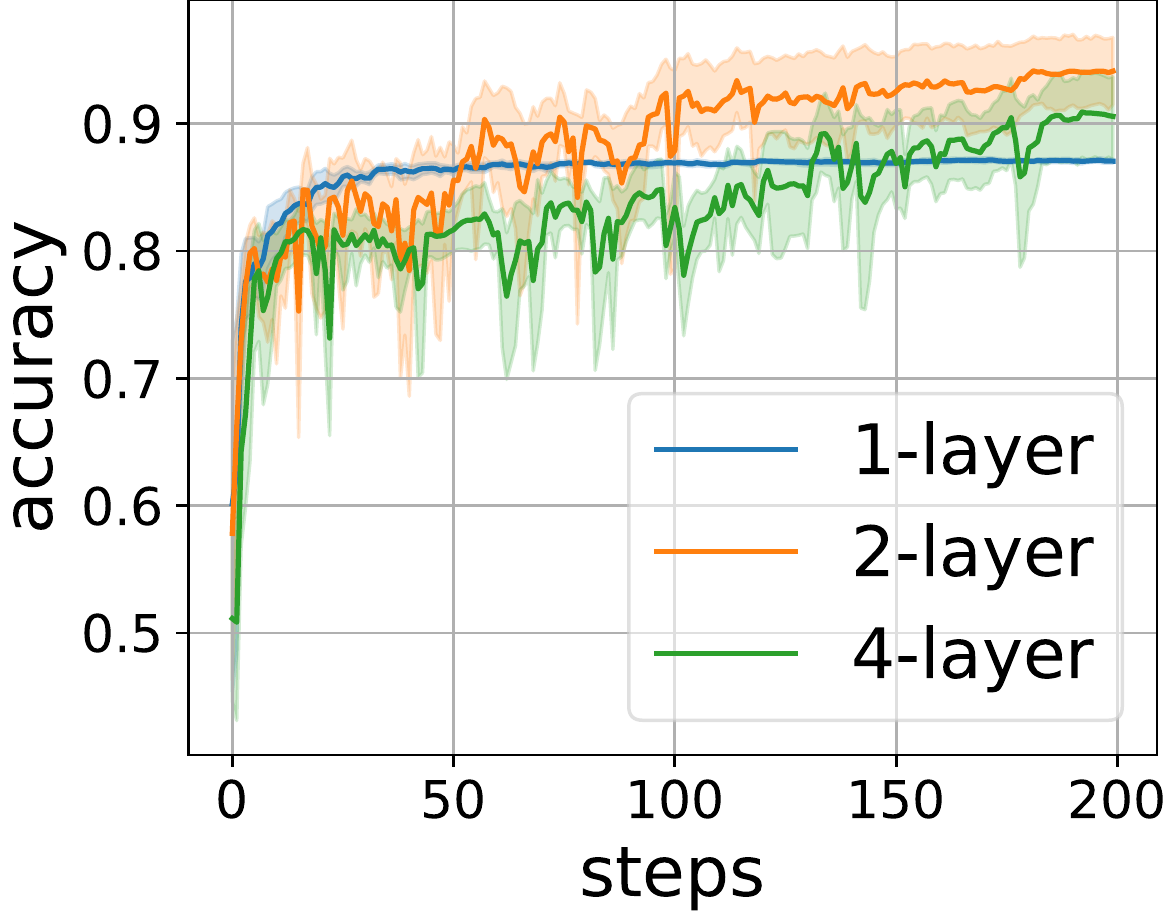}}\\a)
\end{minipage}
\hfill
\begin{minipage}[h]{.24\linewidth}
\center{\includegraphics[width=.94\linewidth]{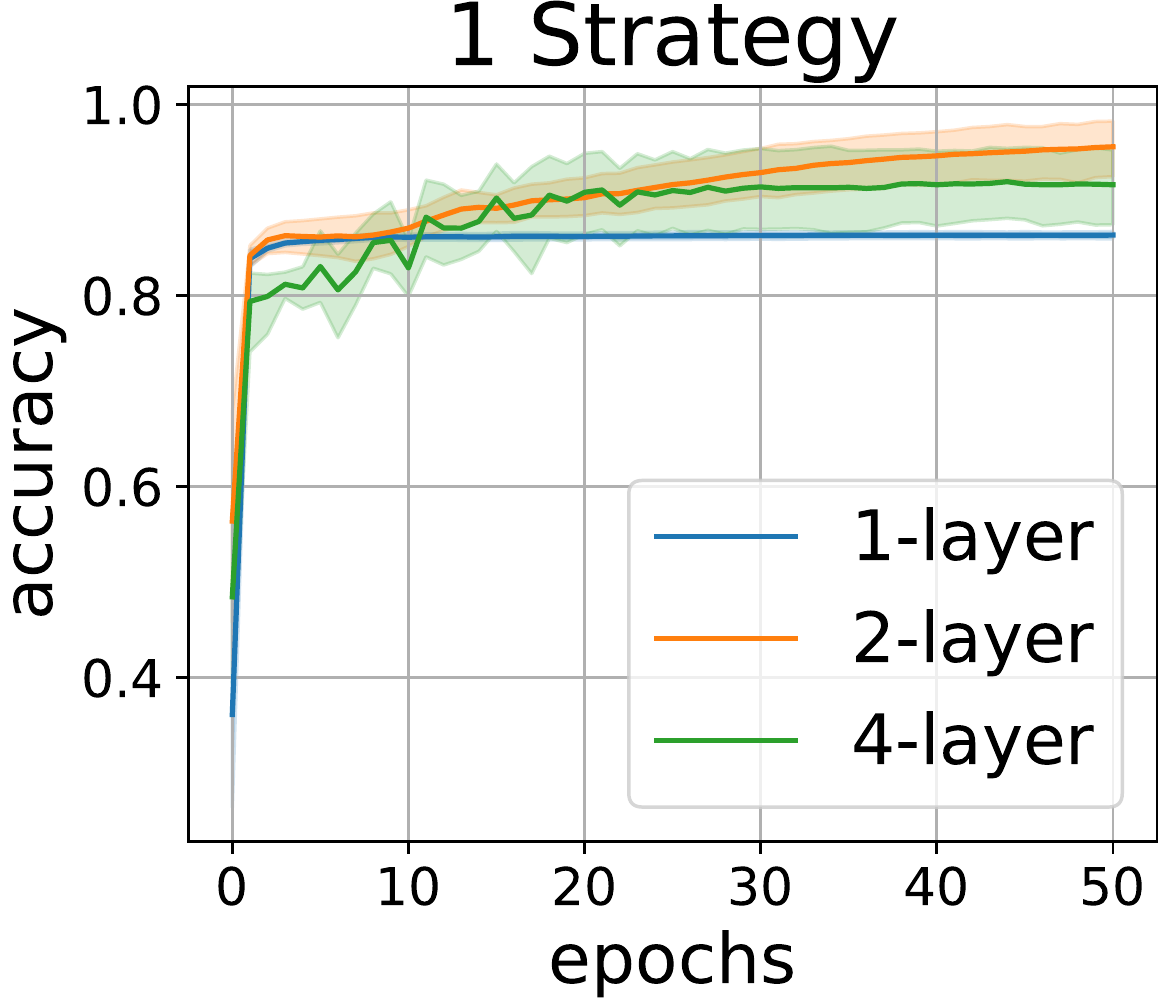}}\\b)
\end{minipage}
\hfill
\begin{minipage}[h]{.24\linewidth}
\center{\includegraphics[width=.94\linewidth]{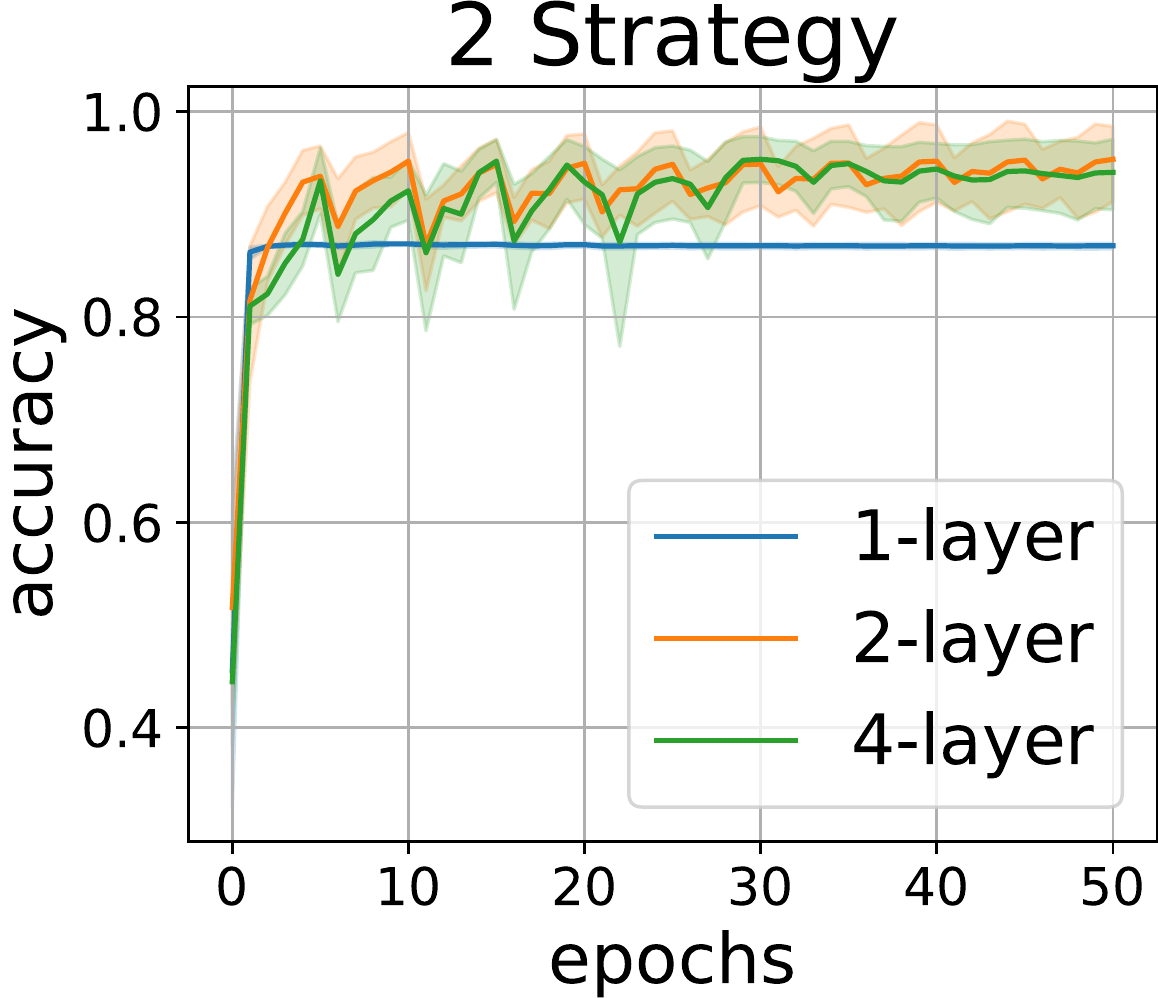}}\\c)
\end{minipage}
\hfill
\begin{minipage}[h]{.24\linewidth}
\center{\includegraphics[width=.94\linewidth]{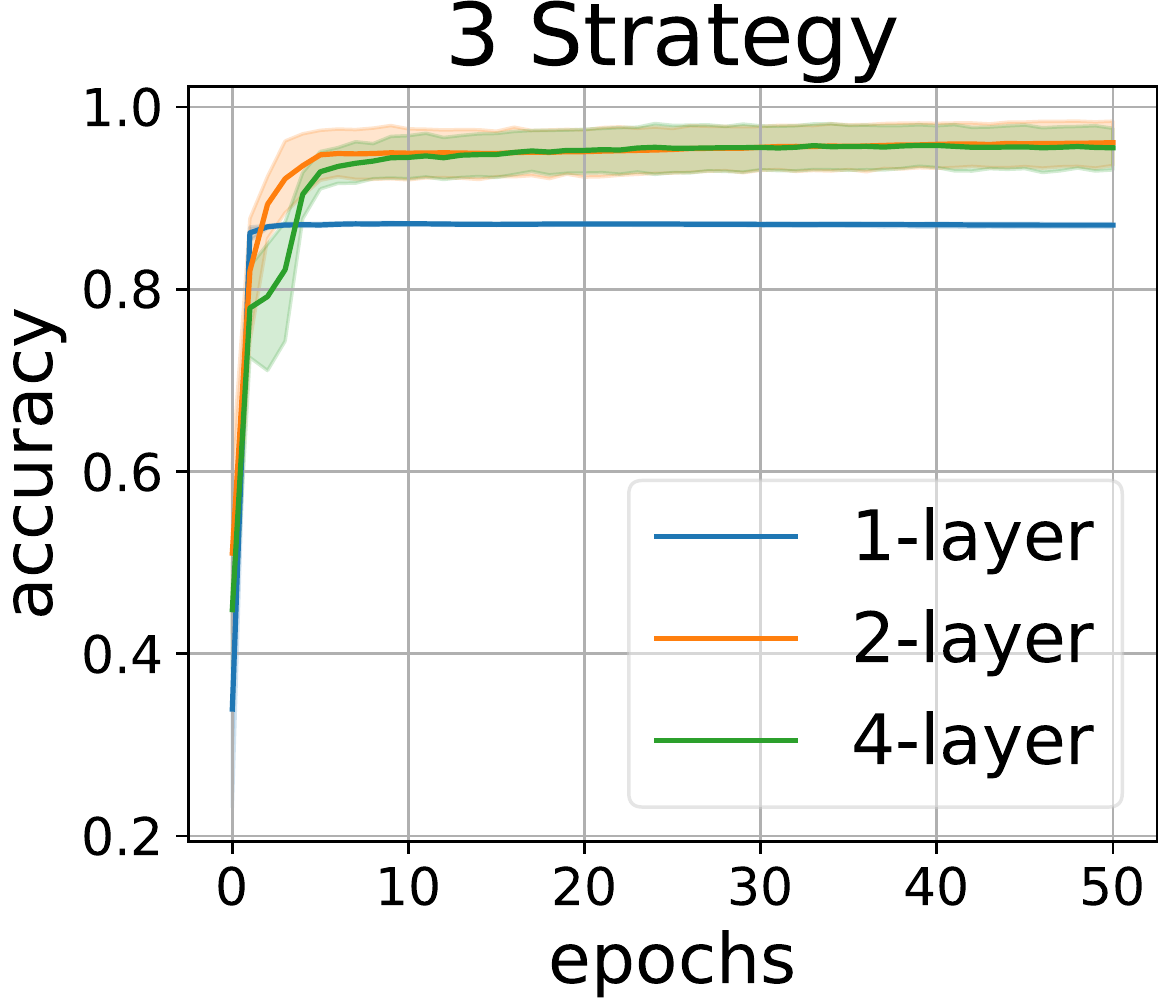}}\\d)
\end{minipage}
\vfill
\begin{minipage}[h]{.24\linewidth}
\center{\includegraphics[width=.94\linewidth]{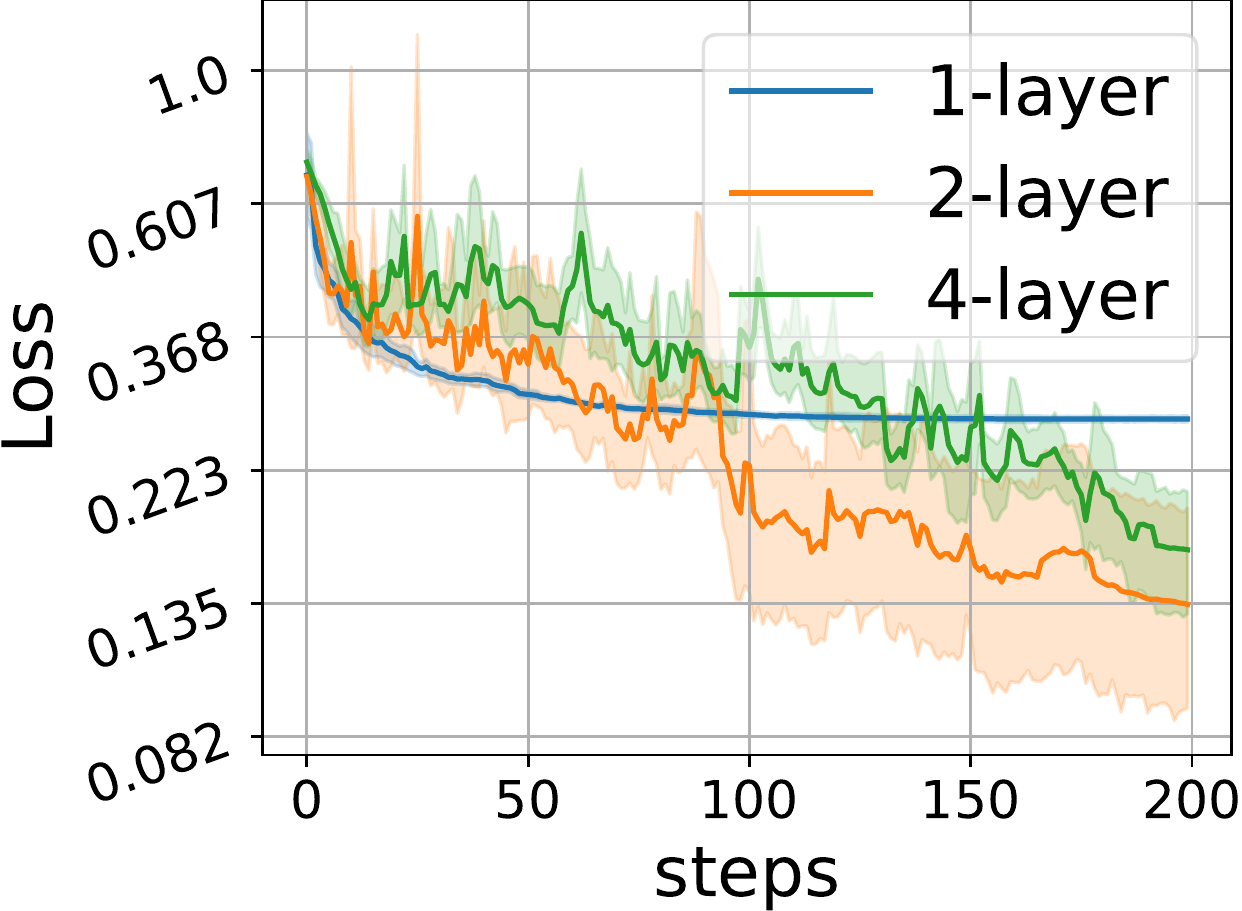}}\\e)
\end{minipage}
\hfill
\begin{minipage}[h]{.24\linewidth}
\center{\includegraphics[width=.94\linewidth]{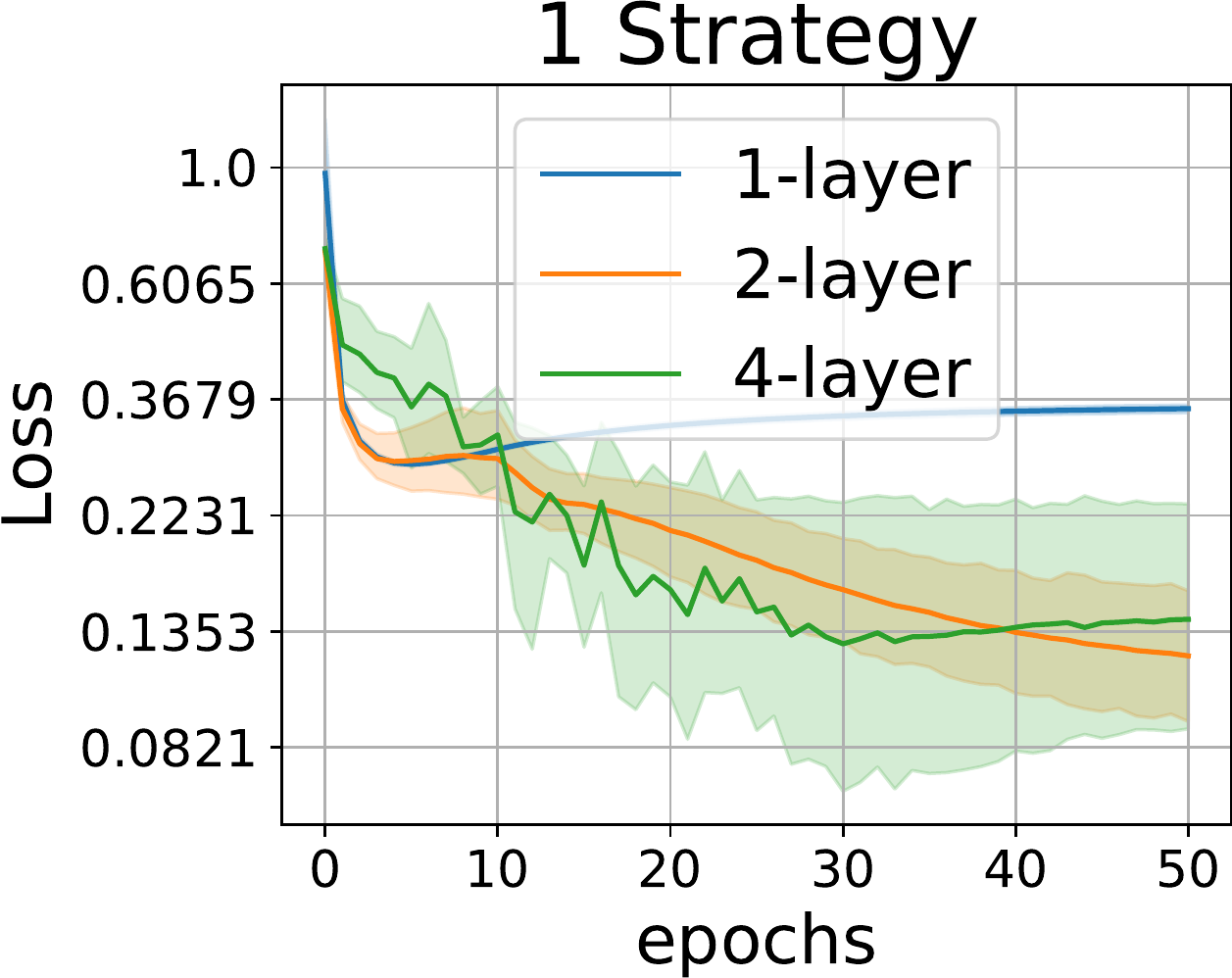}}\\f)
\end{minipage}
\hfill
\begin{minipage}[h]{.24\linewidth}
\center{\includegraphics[width=.94\linewidth]{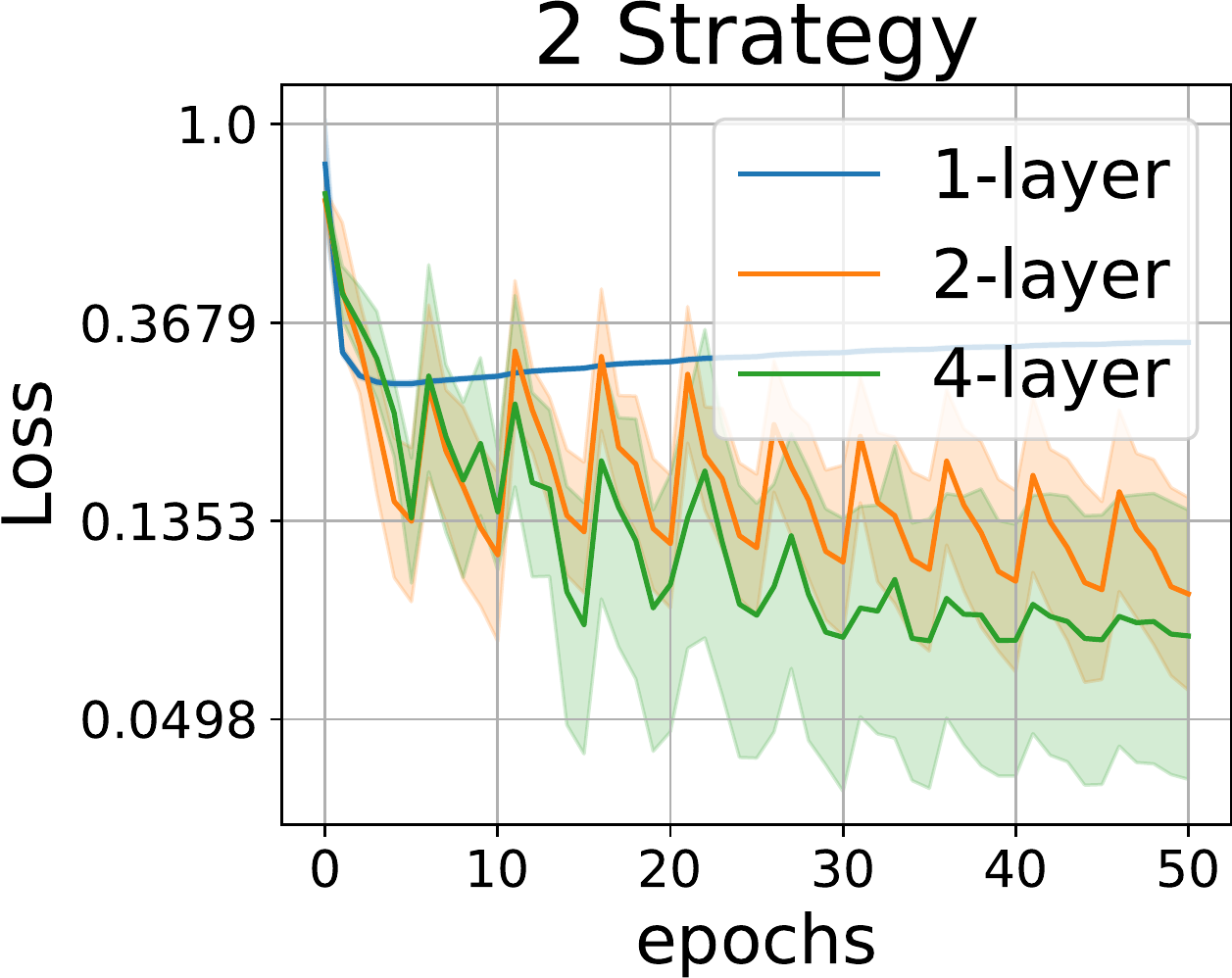}}\\g)
\end{minipage}
\hfill
\begin{minipage}[h]{.24\linewidth}
\center{\includegraphics[width=.94\linewidth]{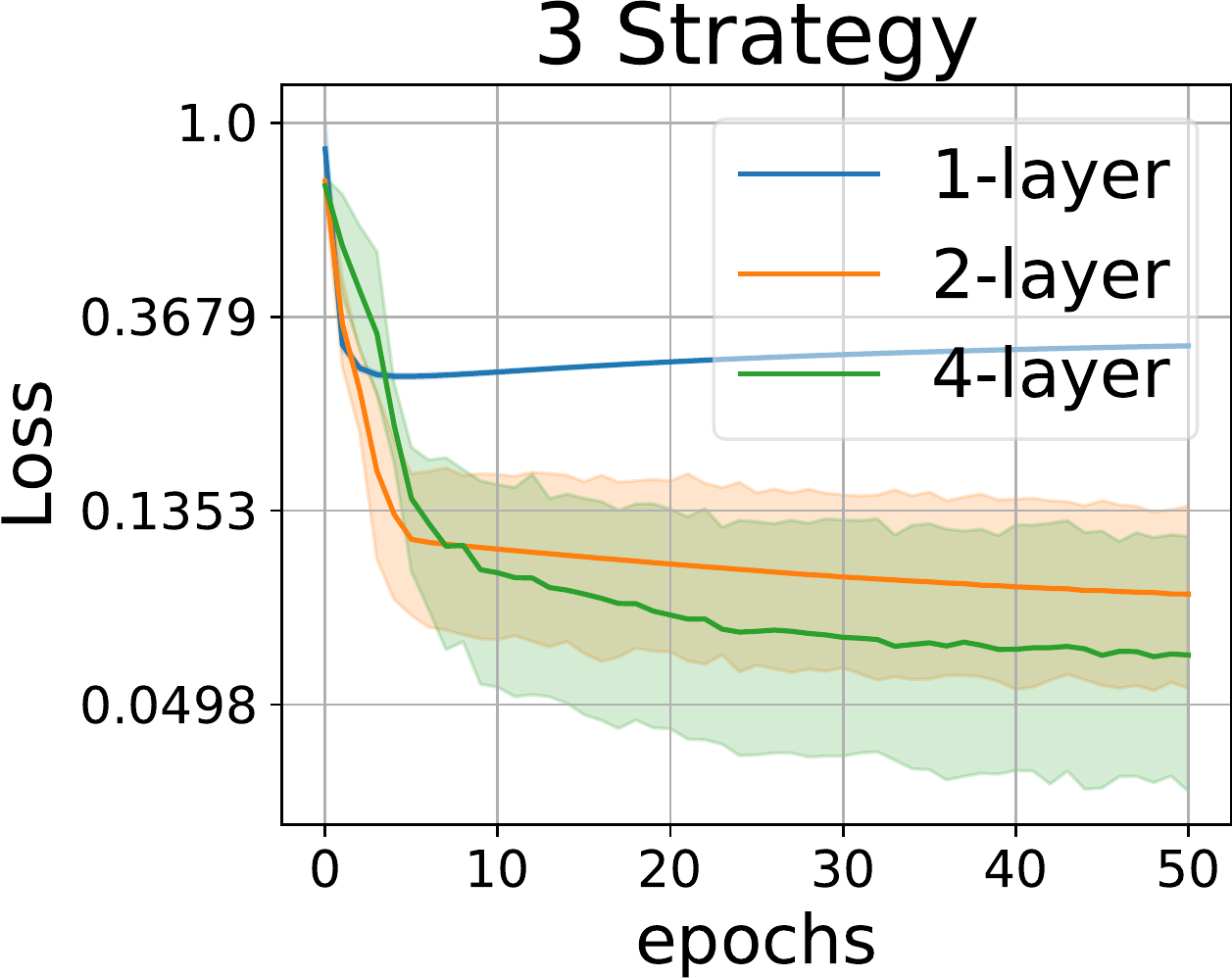}}\\h)
\end{minipage}
\caption{The convergence of the accuracy and logarithmic loss functions on the test part for various strategies of increasing the number of training steps.}
\label{ris8}
\end{figure}

Figures~\ref{ris2}.d and~\ref{ris8} (a, e) show the convergence of both training procedures. Note that the number of steps is not enough to train the 2-layer and 4-layer models on the distilled data. Therefore, we assume that quality will increase if the number of iterations is greater.

\begin{figure}[h!]
\begin{minipage}[h]{\linewidth}
\center{\includegraphics[width=.8\linewidth]{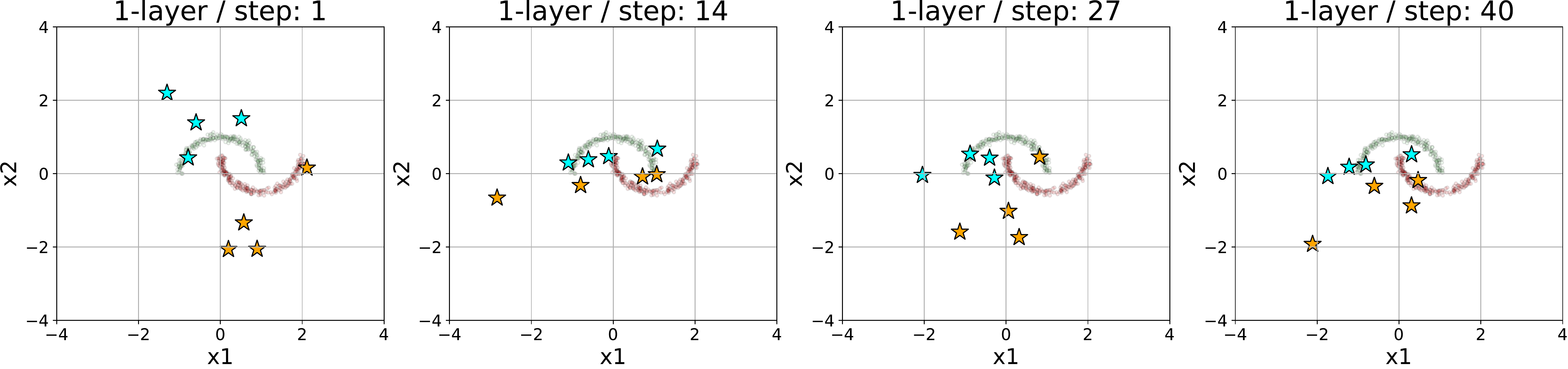}}
\end{minipage}
\hfill
\begin{minipage}[h]{\linewidth}
\center{\includegraphics[width=.8\linewidth]{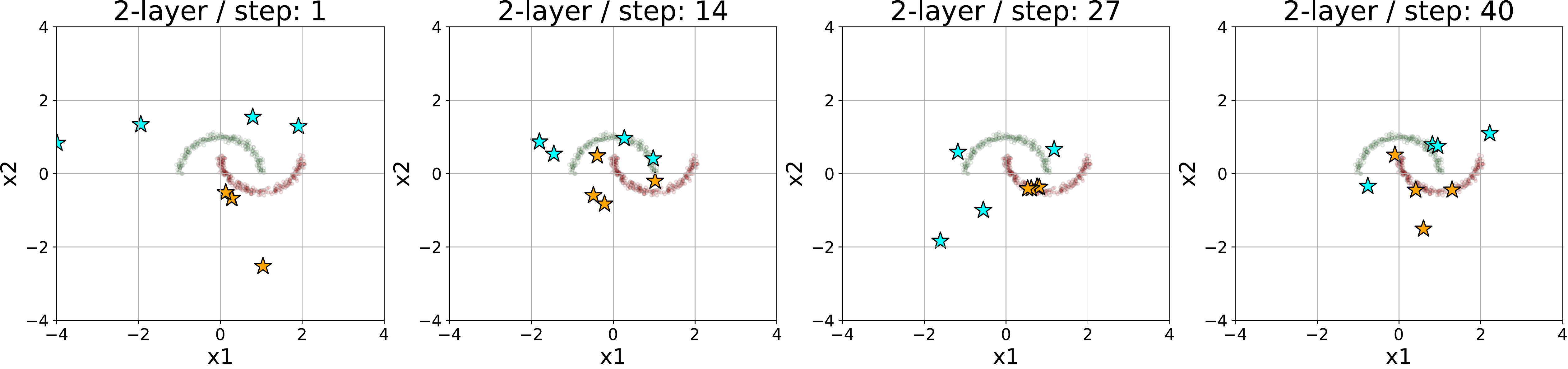}}
\end{minipage}
\hfill
\begin{minipage}[h]{\linewidth}
\center{\includegraphics[width=.8\linewidth]{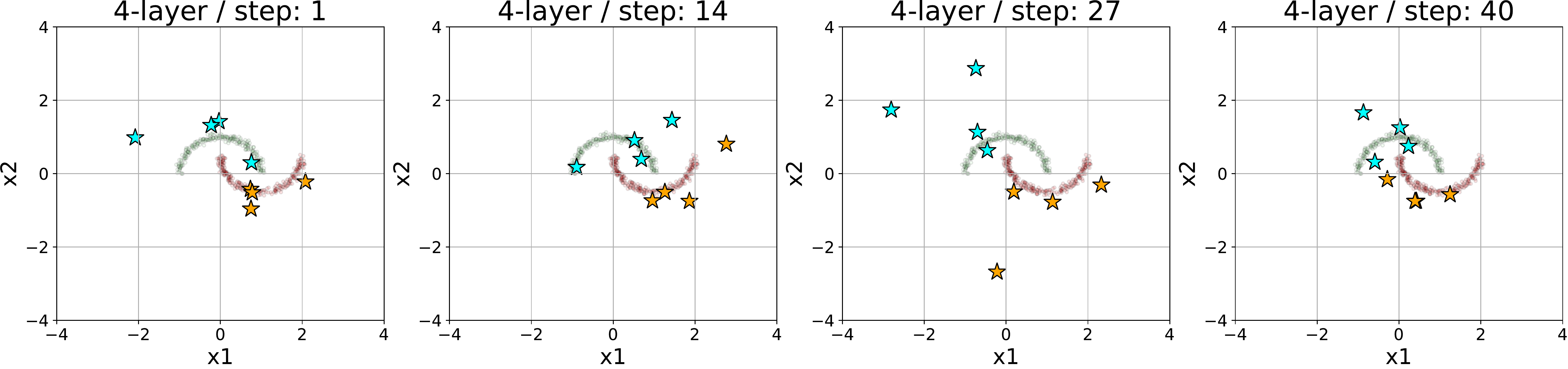}}
\end{minipage}
\caption{Synthetic objects of different internal steps (1st, 14th, 27th and last) that were used to train median-quality models.}
\label{ris6}
\end{figure}

Figure~\ref{ris6} shows synthetic objects from different internal steps for datasets used to achieve median quality. Note that there is no strong similarity between distilled and original objects, as it was in the experiments with images~\cite{l1}. Also, it seems that objects of data distilled for the linear model are often located along the decision boundary. Thus we can assume that the data can overfit for a specific architecture which can be a barrier to train other architectures on such data.

Figure~\ref{ris3} shows the boundaries of the decision rules for median-quality models trained on the original (a, c) and distilled (b, d) data. The progress of the 2-layer architecture is clear: using distilled data models can build more complex decision rules, which bring them closer to bigger models.

\subsection{The Problem of a Small Number of Epochs}
Due to memory and time complexity, it is undesirable to use a large number of internal epochs in the distillation procedure. Fixation of a small number of steps causes the aforementioned problem: it doesn't allow training procedure on distilled data to converge. To overcome this problem, we explore three strategies for artificially increasing the number of epochs when training new models.
\begin{figure}[h!]
\begin{minipage}[h]{\linewidth}
\begin{minipage}[h]{.24\linewidth}
\center{\includegraphics[width=.9\linewidth]{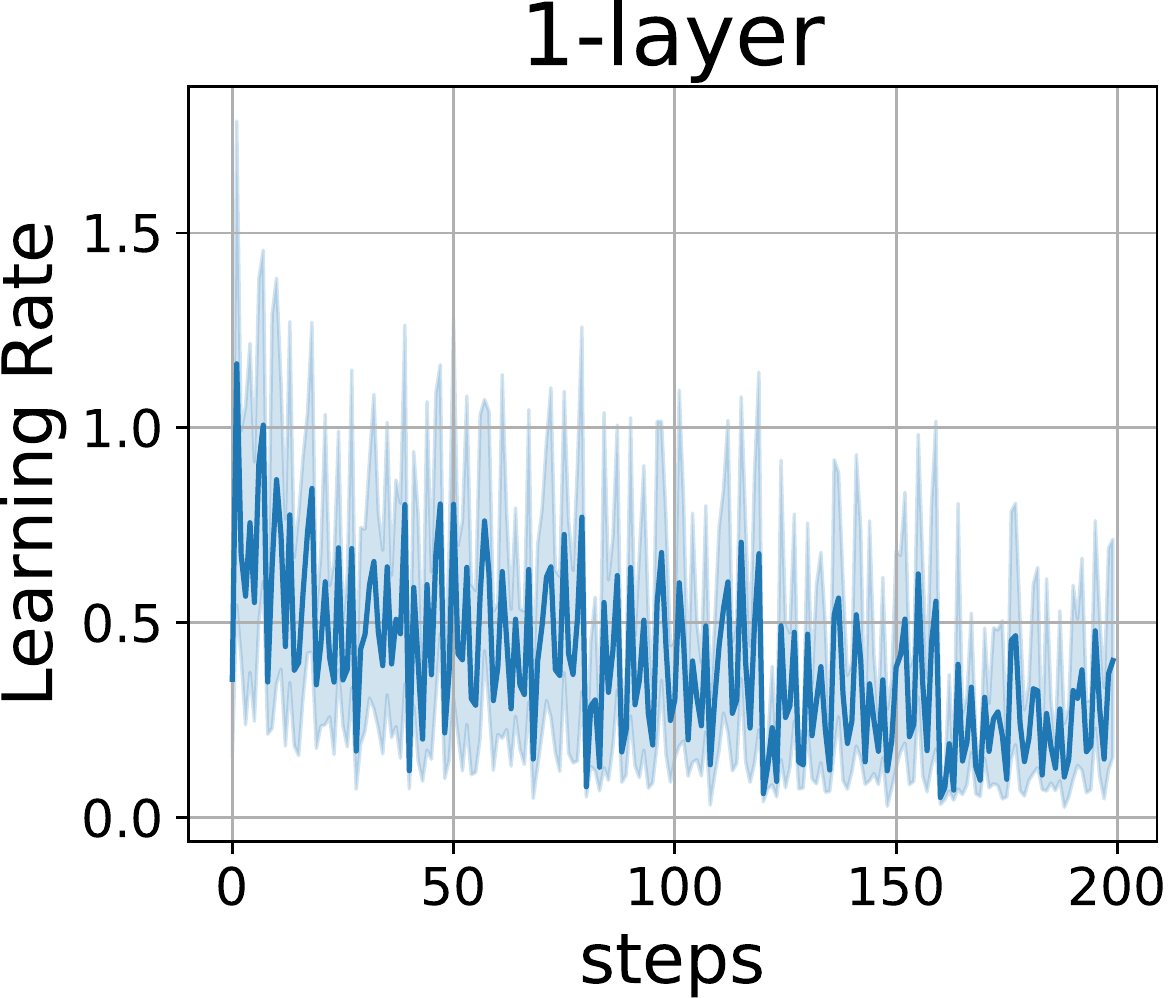}}\\a)
\end{minipage}
\hfill
\begin{minipage}[h]{.24\linewidth}
\center{\includegraphics[width=.9\linewidth]{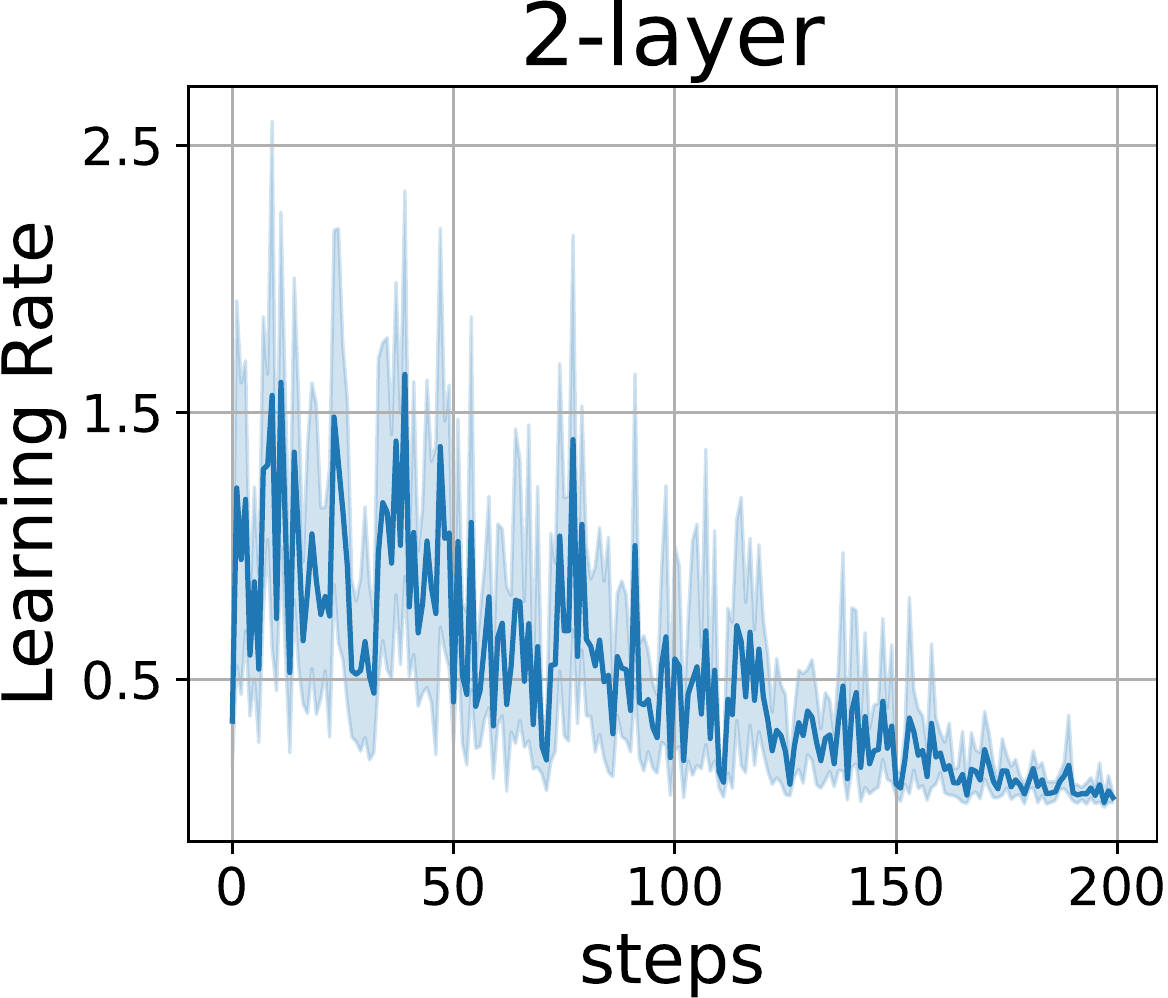}}\\b)
\end{minipage}
\hfill
\begin{minipage}[h]{.24\linewidth}
\center{\includegraphics[width=.9\linewidth]{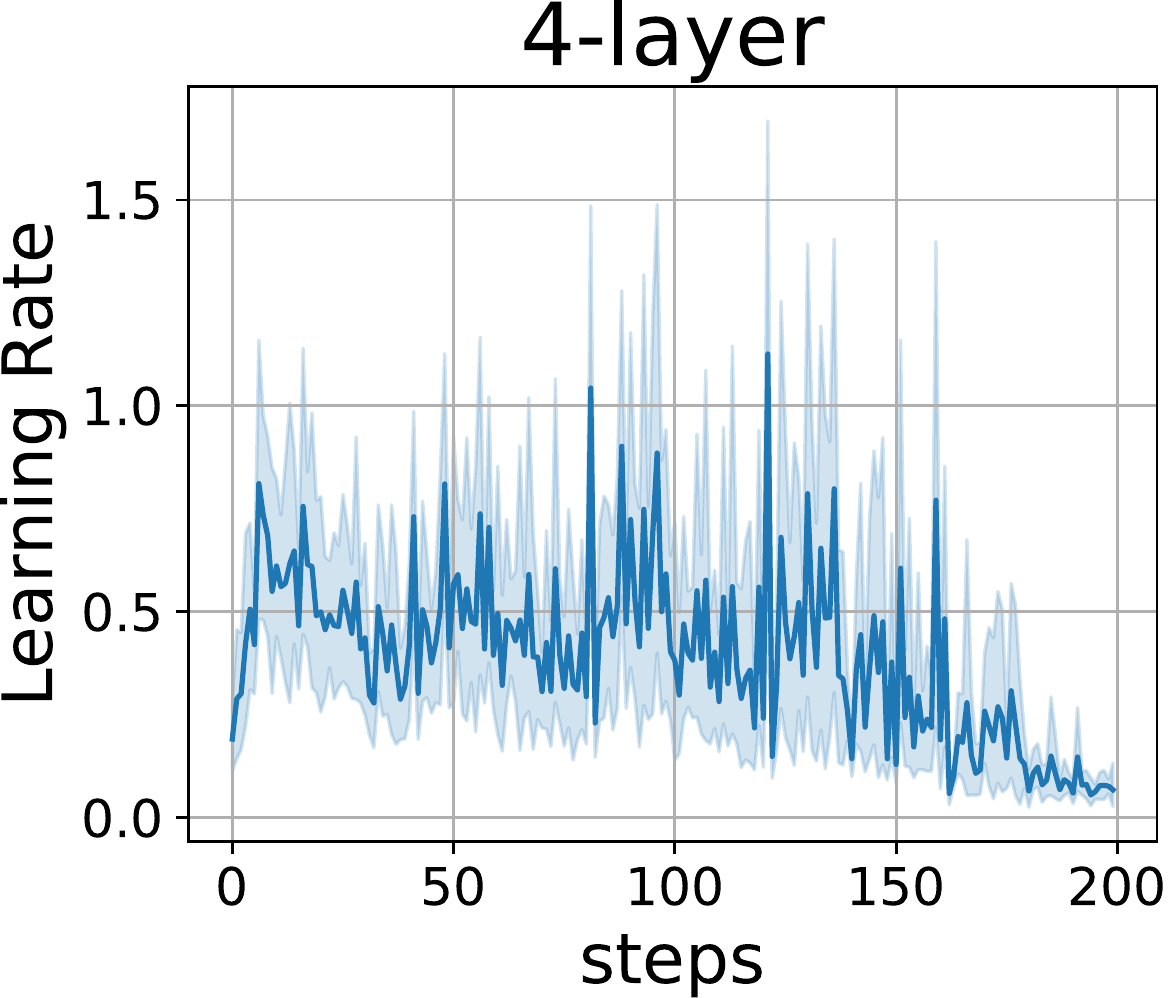}}\\c)
\end{minipage}
\hfill
\begin{minipage}[h]{.24\linewidth}
\center{\includegraphics[width=.9\linewidth]{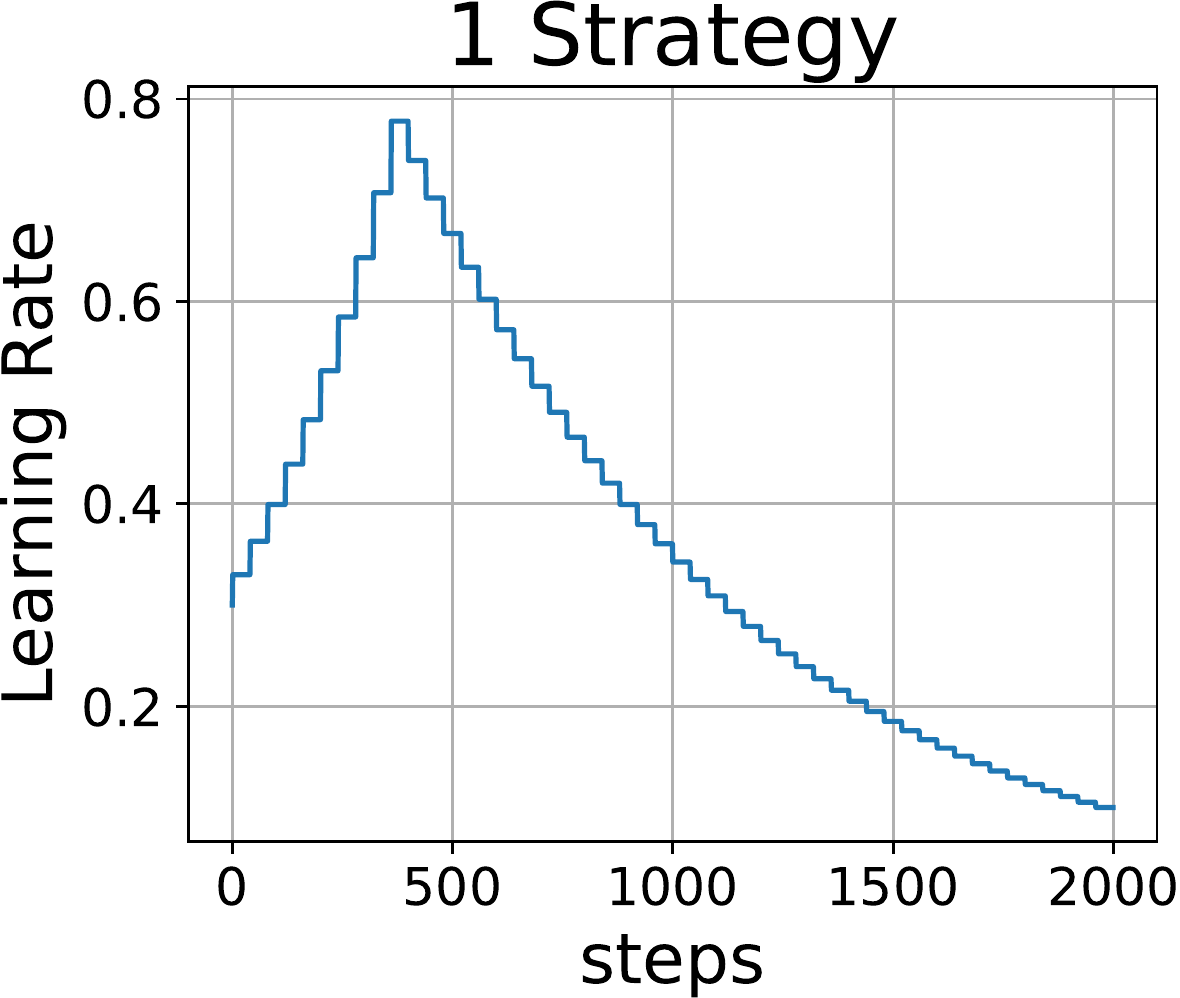}}\\d)
\end{minipage}
\hfill
\end{minipage}
\caption{Learning rates: a-c) obtained by distillation and averaged for 10 different initializations; d) used in the first strategy.}
\label{ris7}
\end{figure}

\begin{figure}[h!]
\begin{minipage}[h]{.32\linewidth}
\center{\includegraphics[width=.7\linewidth]{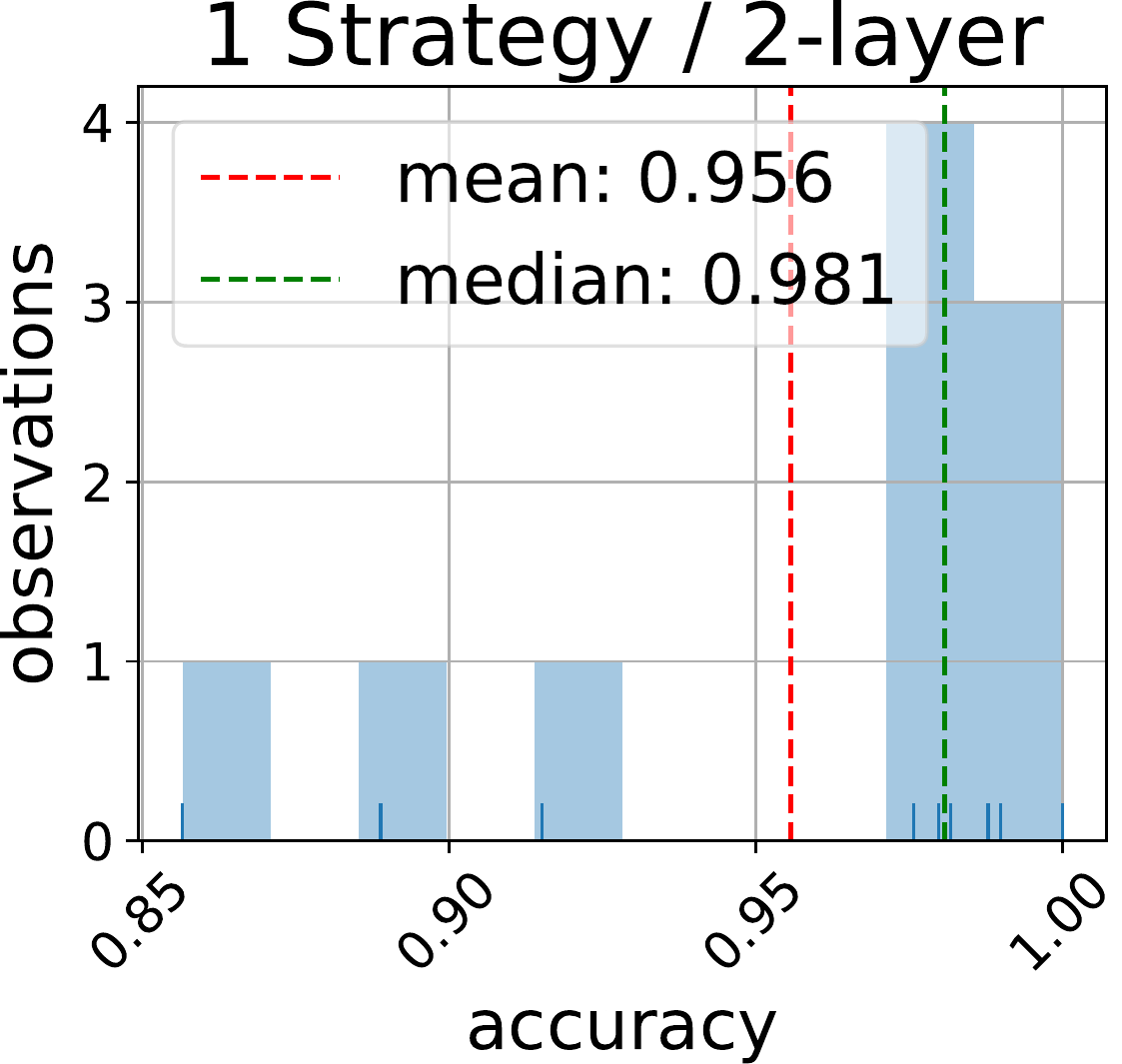}}\\a)
\end{minipage}
\hfill
\begin{minipage}[h]{.32\linewidth}
\center{\includegraphics[width=.7\linewidth]{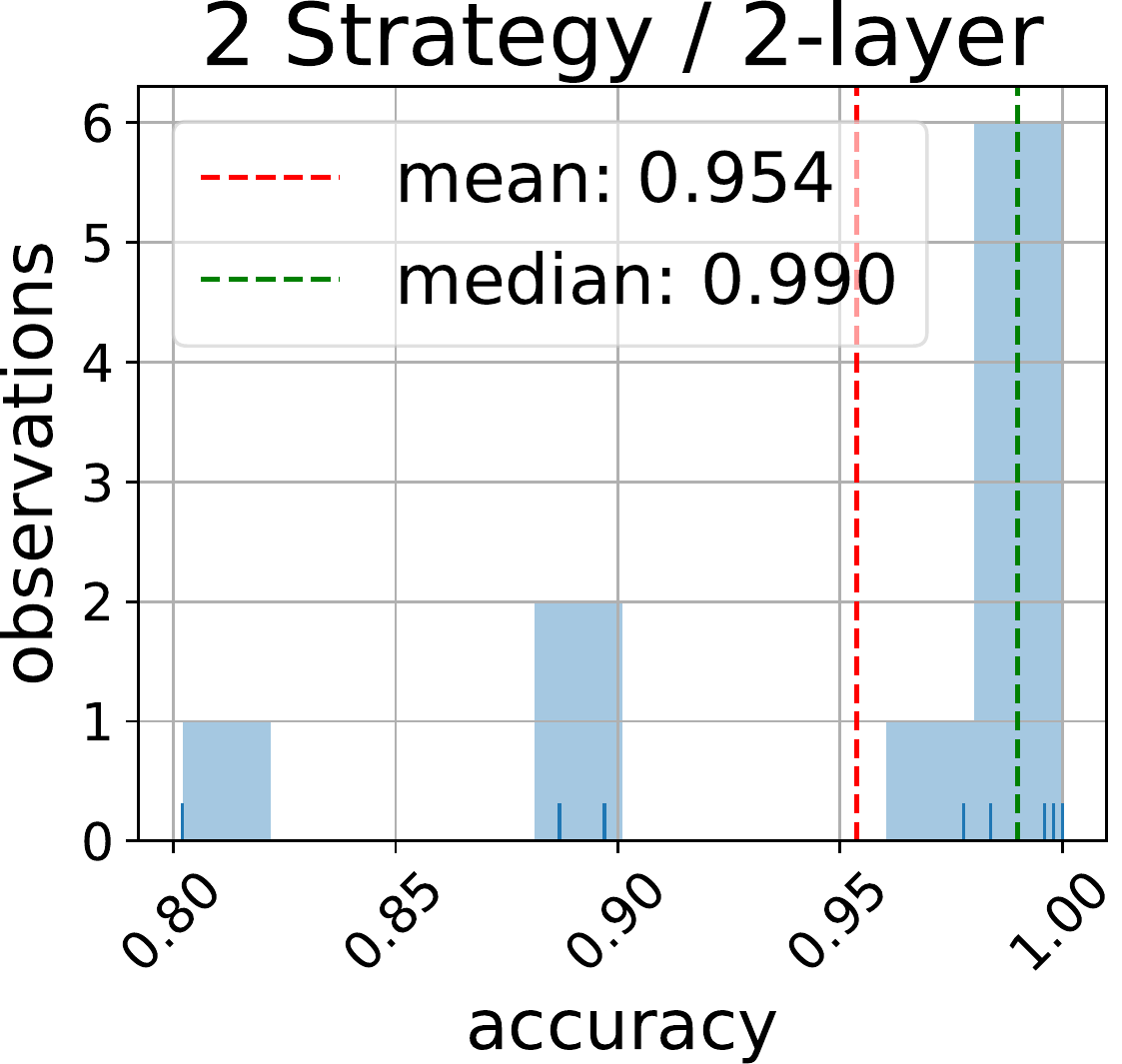}}\\b)
\end{minipage}
\hfill
\begin{minipage}[h]{.32\linewidth}
\center{\includegraphics[width=.7\linewidth]{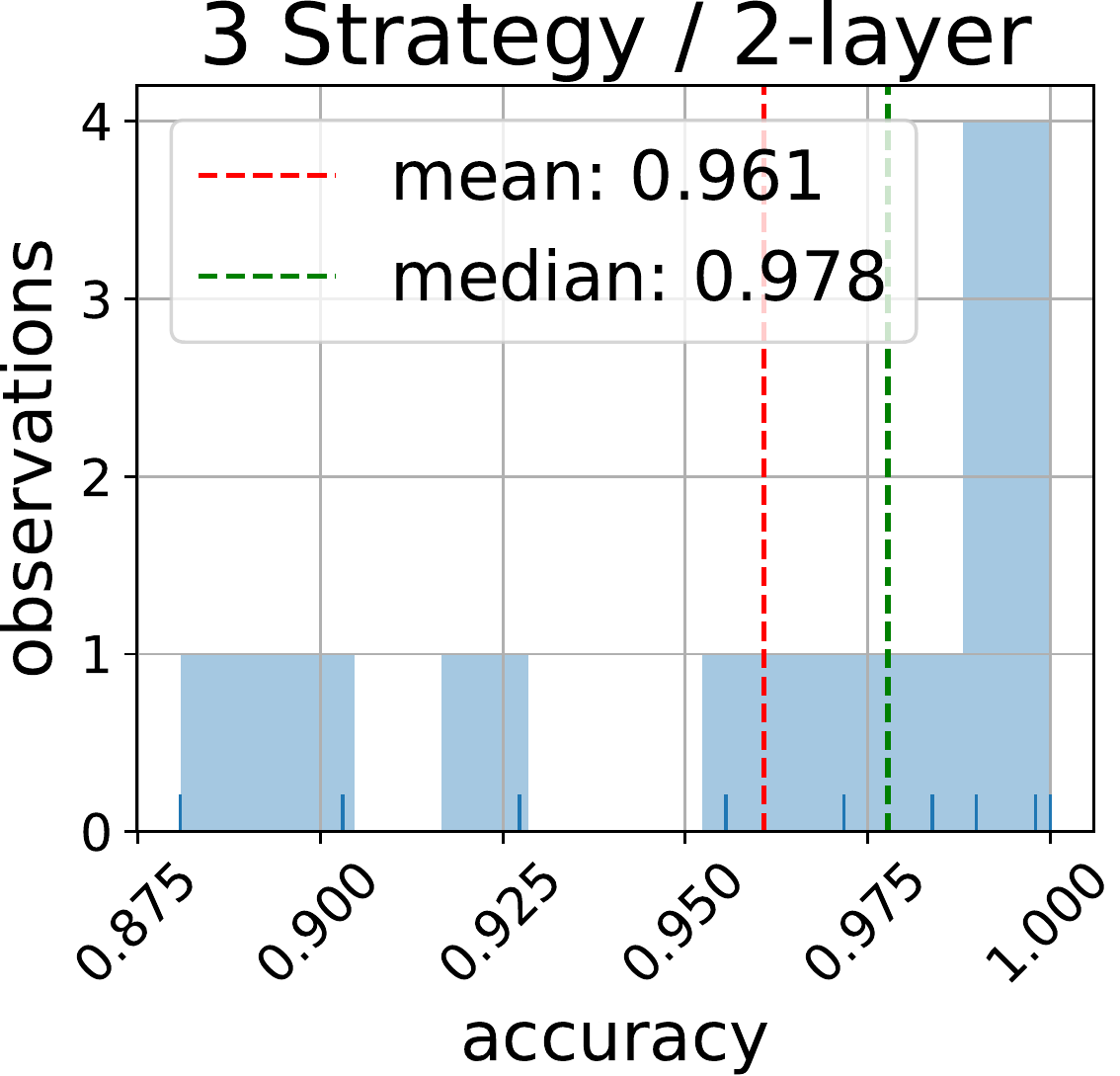}}\\c)
\end{minipage}
\vfill
\begin{minipage}[h]{.32\linewidth}
\center{\includegraphics[width=.7\linewidth]{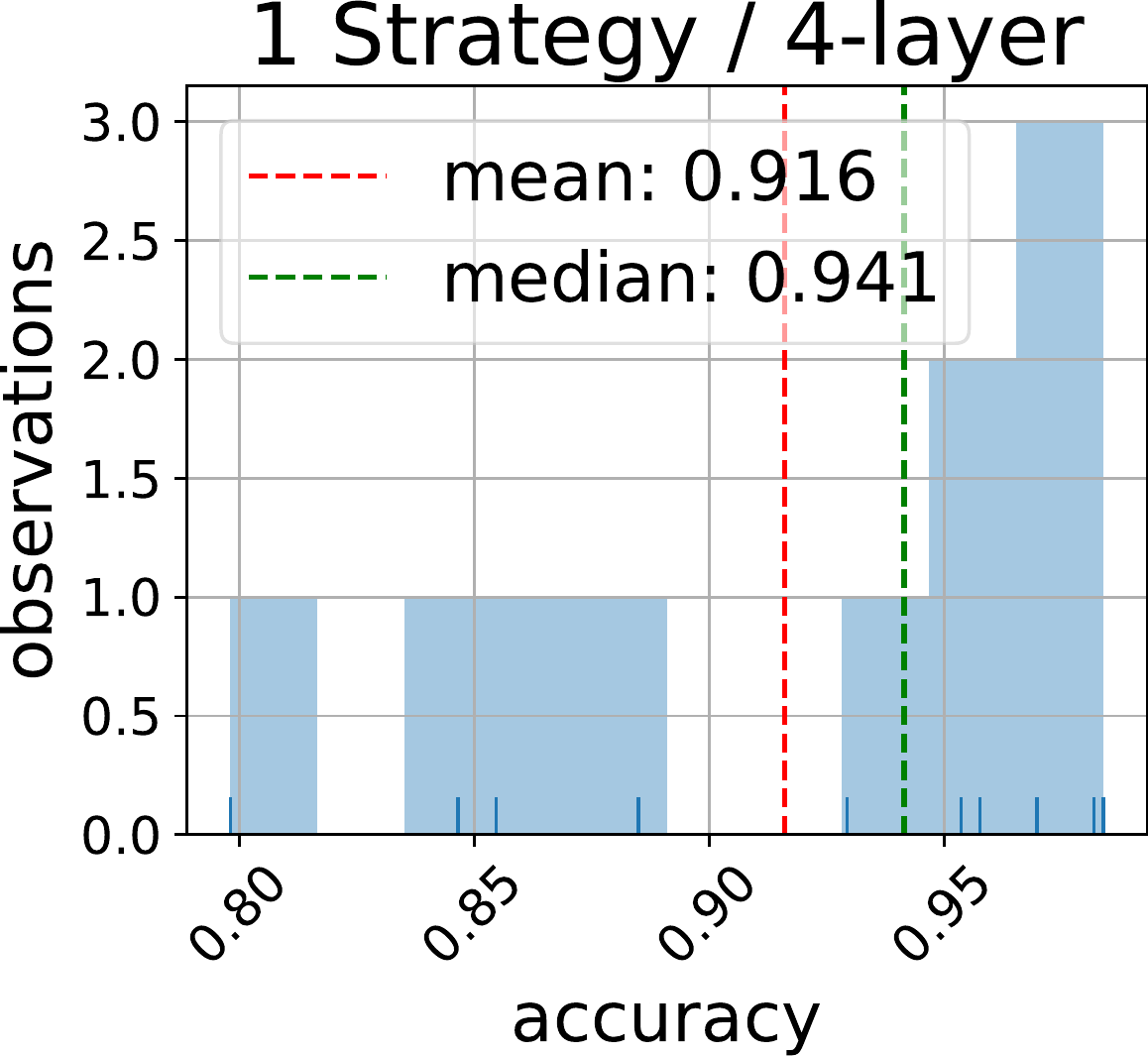}}\\d)
\end{minipage}
\hfill
\begin{minipage}[h]{.32\linewidth}
\center{\includegraphics[width=.7\linewidth]{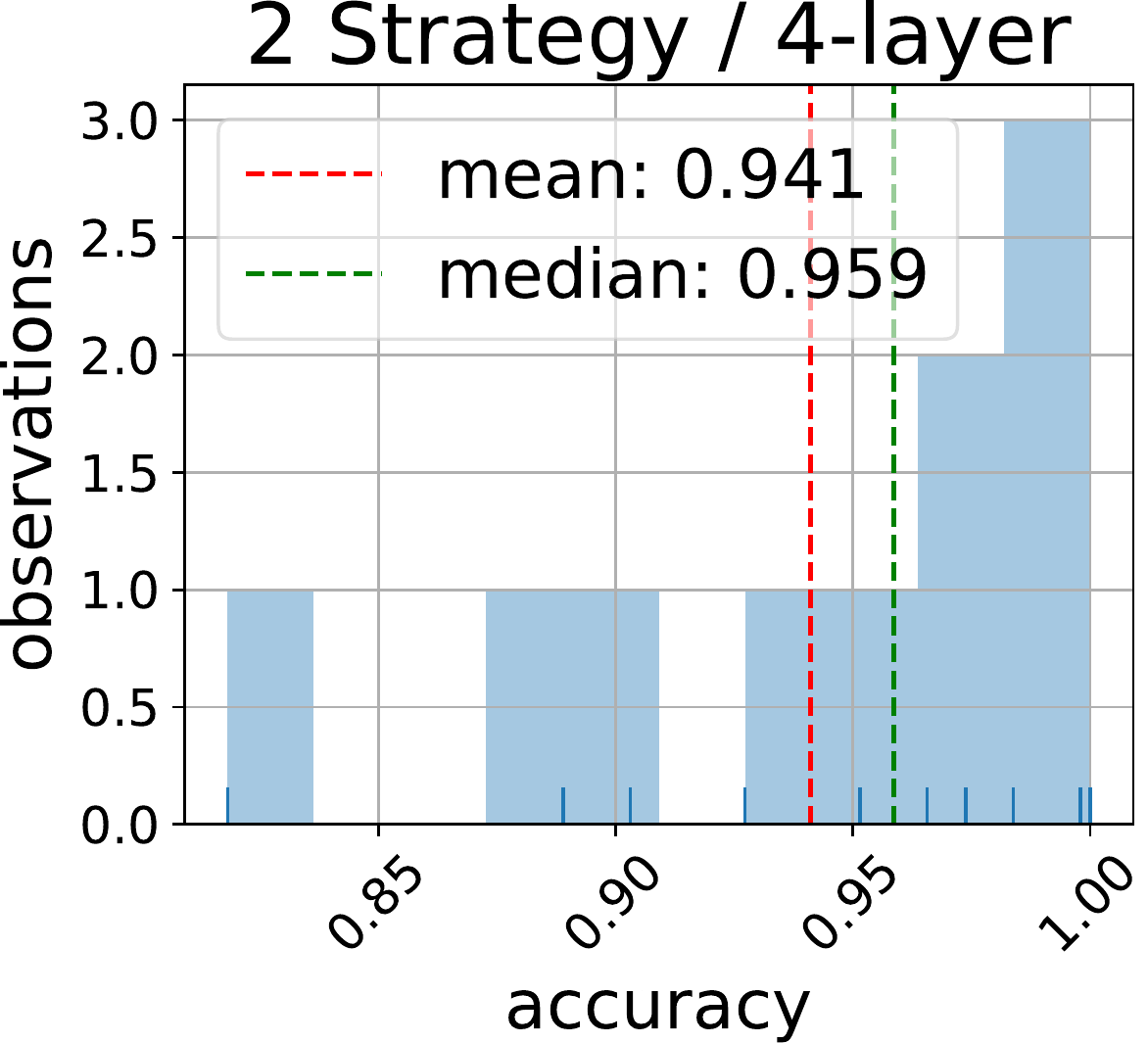}}\\e)
\end{minipage}
\hfill
\begin{minipage}[h]{.32\linewidth}
\center{\includegraphics[width=.7\linewidth]{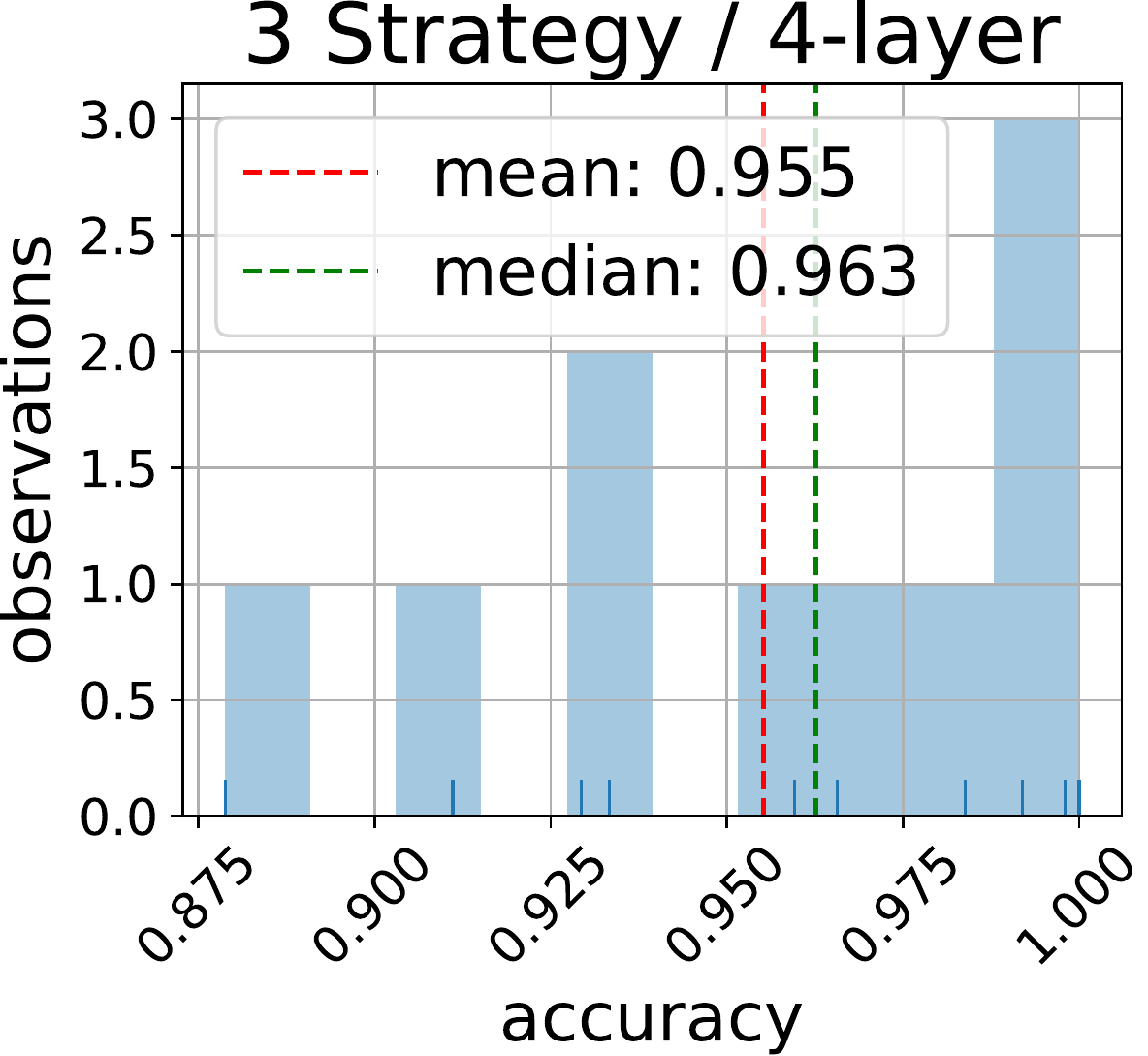}}\\f)
\end{minipage}
\caption{Distribution of accuracy for different initializations, architectures and strategies of increasing the number of training steps.}
\label{ris9}
\end{figure}

\noindent Figure~\ref{ris7} (a, b, c) shows synthetic learning rates. Note that there are strong fluctuations from iteration to iteration, so it seems difficult to just replace such a complex scheme with a universal one. Nevertheless, we show that a standard strategy (see Fig.~\ref{ris7} d) can improve the quality of models trained on distilled data (see Fig.~\ref{ris5},~\ref{ris9} a, d, Table~\ref{table1}). This strategy increases step sizes in 1.1 times at the beginning of each new epoch and then similarly decreases them in 0.95 times. Note that the new strategy allowed non-linear models to converge (see Fig.~\ref{ris8} b, f) even without using synthetic learning rate.

Other strategies use synthetic learning rates. The second strategy repeats synthetic epochs multiplied on a coefficient. Thus the synthetic learning rate stays unchanged for the first 5 epochs. For the next 5 epochs, each learning rate is multiplied on 0.98 and repeated again. Then, to get 5 more epochs, we again multiply each learning rate but on $ 0.98 \cdot 0.98 = 0.9604 $, and so on we repeat this 10 times. Note that in contrast to the smooth convergence of the previous strategy (see Fig.~\ref{ris8} b, f), the new one has strong fluctuations during training (see Fig.~\ref{ris8} c, g), but outperforms the previous one (see Fig.~\ref{ris9} b, e).

The third strategy attempts to correct the inaccurate connection of the epochs of the previous strategy. Instead of cyclically repeating all five epochs, only the last is repeated. Note that the training procedure convergence has indeed become smoother (see Fig.~\ref{ris8} d, h) and models reach the higher quality (see Fig.~\ref{ris9} c, f, Table~\ref{table1}).

\section{Data Generalization to Different Architectures}
One of the possible practical applications of the data distillation is the fast training of a large number of different architectures with different initializations to acceptable quality. Therefore, it is important for the synthetic data to be well generalized to all variations of networks. To examine this issue, we tried to train models of different architectures on each distilled dataset. 

Table~\ref{table1} (first subtable) depicts the results. Note that the synthetic data generally acts acceptable for architectures simpler than used in distillation procedure. The worst result was obtained when training a 4-layer model on the data distilled for a 2-layer. Since the data distilled for 2-layer models doesn't seem much different from the data distilled for 4-layer models, we assume that the problem can be caused by synthetic learning rates. So it makes sense to try each of the three strategies described in the previous section. Next subtables in Table~\ref{table1} show the result: it seems that for all three cases there are some improvements in the quality. The most significant changes touched the worst case. Note that using the strategy without distilled learning rates helped to get much better quality, even higher than when training on data distilled specifically for this architecture.

\begin{table}[h!]
 \centering
 \begin{tabular}{l|c|c|c}
\multirow{2}{*}{Data Models} & \multicolumn{3}{c}{Test Models}\\\cline{2-4}
 & 1-layer & 2-layer & 4-layer\\\hline
original & $0.766\pm 0.089$ & $0.877 \pm 0.005$ & \boldmath{$0.995 \pm 0.015$}\\\hline
1-layer & \boldmath{$0.871\pm 0.003$} & $0.869 \pm 0.004$ & $0.864 \pm 0.006$\\\hline
2-layer & $0.808 \pm 0.014$ & \boldmath{$0.941 \pm 0.043$} & $0.691 \pm 0.182$\\\hline
4-layer  &$0.825 \pm 0.014$ & $0.879 \pm 0.013$ & $0.906 \pm 0.054$\\
 \multicolumn{4}{c}{}\\
Strategy1 + 1-layer & \boldmath{$0.863 \pm 0.006$} & $0.860 \pm 0.008 $ & $0.860 \pm 0.010$ \\\hline
Strategy1 + 2-layer & $0.808 \pm 0.010$ & \boldmath{$0.956 \pm 0.047$} & \boldmath{$0.985 \pm 0.015$} \\\hline
Strategy1 + 4-layer & $0.818 \pm 0.012$ & $0.911 \pm 0.059$ & $0.916 \pm 0.062$ \\
\multicolumn{4}{c}{}\\
Strategy2 + 1-layer & \boldmath{$0.869 \pm 0.004$} & $0.867 \pm 0.006$ & $0.865 \pm 0.005$ \\\hline
Strategy2 + 2-layer & $0.804 \pm 0.012$ & \boldmath{$0.954 \pm 0.065$} & $0.672 \pm 0.229$ \\\hline
Strategy2 + 4-layer & $0.827 \pm 0.017$ & $0.937 \pm 0.035$ & \boldmath{$0.941 \pm 0.055$}\\
\multicolumn{4}{c}{}\\
Strategy3 + 1-layer & \boldmath{$0.870 \pm 0.003$} & $0.866 \pm 0.006$ & $0.863 \pm 0.008$ \\\hline
Strategy3 + 2-layer  & $0.807 \pm 0.011$ & \boldmath{$0.961 \pm 0.041$} & $0.834 \pm 0.210$ \\\hline
Strategy3 + 4-layer  & $0.835 \pm 0.016$ & $0.910 \pm 0.041$ & \boldmath{$0.955 \pm 0.039$}
 \end{tabular}
 \caption{Mean and standard deviation of accuracy on the test part for different sets of synthetic data and models. Bold font indicates the biggest value in the column.}\label{table1}
 \end{table}

\begin{figure}[h!]
\begin{minipage}[h]{\linewidth}
\center{\includegraphics[width=.9\linewidth]{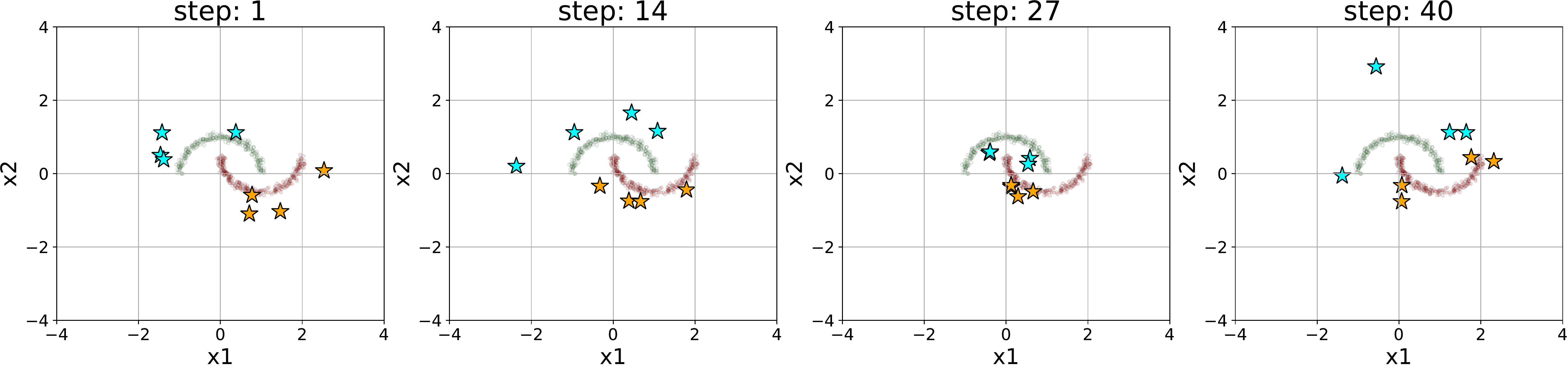}}\\
\end{minipage}
\caption{Synthetic objects of different internal steps (1st, 14th, 27th and last) that were used to train median-quality models. Data distilled using all three architectures.}
\label{ris10}
\end{figure}

The natural assumption is that data distillation using all three architectures can lead to better results. to do so we select $m = 3$, but instead of using three internal models with the same architecture, we use three different ones: 1-layer, 2-layer and 4-layer. Table~\ref{table2} shows the results: for all non-linear architectures, we were able to reach higher accuracy using undistilled learning rates. Note that new synthetic data (see Fig.~\ref{ris10}) looks similar to the data from previous experiments (see Fig.~\ref{ris6}).

\begin{table}[h!]
 \centering
 \begin{tabular}{l|c|c|c}
\multirow{2}{*}{Data models} & \multicolumn{3}{c}{Test Models}\\\cline{2-4}
 & 1-layer & 2-layer & 4-layer \\\hline
 raw steps & $0.859 \pm 0.005$ & $0.881 \pm 0.004$ & $0.867 \pm 0.122$\\\hline
 strategy 1 & $0.851 \pm 0.007$ & \boldmath{$0.970 \pm 0.028$} & \boldmath{$0.986 \pm 0.014$}\\\hline
 strategy 2 & \boldmath{$0.862 \pm 0.007$} & $0.941 \pm 0.027$ & $0.984 \pm 0.017$\\\hline
 strategy 3 & $0.858 \pm 0.006$ & $0.897 \pm 0.014$ & $0.965 \pm 0.045$\\\hline
 \end{tabular}
 \caption{Mean and standard deviation of accuracy on the test part for different sets of synthetic data and models. Bold font indicates the biggest value in the column. Data distilled using all three architectures.}\label{table2}
 \end{table} 



\section{Conclusion}
In this work, we examined the distillation of tabular data using the algorithm proposed in~\cite{l1}. We observed that models trained using distilled data can outperform models trained on the whole original data. We show that synthetic objects have some generalizability and can be successfully used in the training of different architectures. In addition, we found that it is sometimes better not to use synthetic learning rates, and explored some strategies to increase the number of training steps. As future work, we plan to change the memory complexity according to the work~\cite{l4}. Also, we want to improve the distilled data generalizing ability using stochastic depth networks~\cite{l5}. Finally, we would like to bring the distribution of synthetic objects closer to the original.

\section{Acknowledgments}
We thank Sergey Ivanov for detailed feedback on the initial version of this paper. This research was performed at the Center for Big Data Storage and Analysis of Lomonosov Moscow State University and was supported by the National Technology Initiative Foundation (13/1251/2018 of December 11, 2018).
%
%
%

\begin{thebibliography}{8}
\bibitem{l1}
Wang, T., Zhu, J., Torralba, A., Efros, A. A.: Dataset Distillation. CoRR; abs/1811.10959 (2018)

\bibitem{l13} Hinton, G., Vinyals, O., Dean, J.: Distilling the Knowledge in a Neural Network. In: NIPS Deep Learning and Representation Learning Workshop. (2015)

\bibitem{l6} Sucholutsky, I., Schonlau, M.: Soft-Label Dataset Distillation and Text Dataset Distillation. CoRR; abs/1910.02551 (2019)

\bibitem{l2} MNIST Handwritten Digit Database, \url{http://yann.lecun.com/exdb/mnist/}. Last accessed 24 Jun 2020.

\bibitem{l3} Lecun, Y., Bottou, L., Bengio, Y., Haffner, P.: Gradient-Based Learning Applied to Document Recognition. In: Proceedings of the IEEE, vol. 86, pp. 2278--2324 (1998)

\bibitem{l9} LeCun, Y., Boser, B., Denker, J. S., Henderson, D., Howard, R. E., Hubbard, W., and Jackel, L. D.: Backpropagation Applied to Handwritten Zip Code RecognitionNeural Computation. Neural Computation 1(4), 541--551 (1989)

\bibitem{l10} Domke, J.: Generic Methods for Optimization-Based Modeling. In: Proceedings of the Fifteenth International Conference on Artificial Intelligence and Statistics, pp. 318--326. PMLR (2012)

\bibitem{l4} Maclaurin, D., Duvenaud, D. and Adams, R.: Gradient-Based Hyperparameter Optimization Through Reversible Learning. CoRR; abs/1502.03492 (2015)

\bibitem{l7} Bengio, Y.: Gradient-Based Optimization of Hyperparameters. Neural Computation 12(8), 1889--1900 (2000)

\bibitem{l8} Baydin, A., Pearlmutter, B.: Automatic Differentiation of Algorithms for Machine Learning. In: Proceedings of the AutoML Workshop at the International Conference on Machine Learning (ICML). Beijing, China, June 21--26 (2014) 

\bibitem{l11} Liu, D. C., Nocedal, J.: On the Limited Memory BFGS Method for Large Scale Optimization. Mathematical Programming 45, 503--528 (1989)

\bibitem{l12} Polyak, B.: Some Methods of Speeding Up the Convergence of Iteration Methods. USSR Computational Mathematics and Mathematical Physics, vol. 4, pp. 1--17 (1964)

\bibitem{l5} Huang, G., Sun, Y., Liu, Z., Sedra, D. and Weinberger, K.: Deep Networks With Stochastic Depth. CoRR; abs/1603.09382 (2016)

\end{thebibliography}
%

\end{document}